\documentclass[10pt,twocolumn,letterpaper]{article}

\usepackage{cvpr}
\usepackage{authblk}
\usepackage{relsize}
\usepackage{times}
\usepackage{epsfig}
\usepackage{graphicx}
\usepackage{amsmath}
\usepackage{amssymb}
\usepackage{paralist}
\usepackage[algo2e]{algorithm2e}
\usepackage{algorithm}
\usepackage{xfrac}
\usepackage{bm}
\usepackage{caption}
\usepackage{subcaption}
\usepackage{float}
\usepackage{color,soul}
\usepackage{xcolor}
\usepackage{mathtools}
\usepackage{multirow}
\usepackage{hhline}
\usepackage{tabu}
\usepackage[titletoc,page]{appendix}
\usepackage[utf8]{inputenc}
\usepackage[algo2e]{algorithm2e}
\usepackage{algorithm}

\usepackage[pagebackref=true,breaklinks=true,letterpaper=true,colorlinks,bookmarks=false]{hyperref}

\cvprfinalcopy % *** Uncomment this line for the final submission

\def\cvprPaperID{} % *** Enter the CVPR Paper ID here

\renewcommand{\vec}[1]{\mathbf{#1}}
\newcommand{\set}[1]{\mathcal{#1}}

\DeclareMathOperator*{\argmax}{arg\,max}

\renewcommand{\eg}{e.g.\ }
\renewcommand{\wrt}{w.r.t.\ }
\newcommand*{\tran}{^{\mkern-1.5mu\mathsf{T}}}

\newcommand*{\inverse}{^{\mkern-1.5mu{-1}}}
\newcommand{\tssdagger}{\textsuperscript{\textdagger}}
\newcommand{\tssstar}{\textsuperscript{*}}

\definecolor{nicegreen}{HTML}{99C000}
\definecolor{nicegreen2}{HTML}{81d41a}
\definecolor{niceyellow}{HTML}{FDCA00}
\definecolor{niceorange}{HTML}{f5a300}
\definecolor{niceblue}{HTML}{0083cc}
\definecolor{nicepurple}{HTML}{a60084}

\begin{document}

%%%%%%%%% TITLE
\title{CONSAC: Robust Multi-Model Fitting by Conditional Sample Consensus }

\renewcommand\Authsep{, }
\renewcommand\Authand{, }
\renewcommand\Authands{, }
\renewcommand\Authfont{\normalsize\relsize{0.9}}
\renewcommand\Affilfont{\normalsize\relsize{0.2}}
\setlength{\affilsep}{.5em} 
\makeatletter
\renewcommand\AB@affilsepx{, \protect\Affilfont}
\makeatother

\author[1]{Florian Kluger}
\author[2]{Eric Brachmann}
\author[1]{Hanno Ackermann}
\author[2]{Carsten Rother}
\author[3]{Michael Ying Yang}
\author[1]{Bodo Rosenhahn}

\affil[1]{Leibniz University Hannover}
\affil[2]{Heidelberg University}
\affil[3]{University of Twente}

\maketitle
% \thispagestyle{empty}

%%%%%%%%% ABSTRACT
\begin{abstract}

% ============= Eric v.3 ==================================================
We present a robust estimator for fitting multiple parametric models of the same form to noisy measurements. 
Applications include finding multiple vanishing points in man-made scenes, fitting planes to architectural imagery, or estimating multiple rigid motions within the same sequence. 
In contrast to previous works, which resorted to hand-crafted search strategies for multiple model detection, we learn the search strategy from data.
A neural network conditioned on previously detected models guides a RANSAC estimator to different subsets of all measurements, thereby finding model instances one after another.
We train our method supervised as well as self-supervised.
For supervised training of the search strategy, we contribute a new dataset for vanishing point estimation. 
Leveraging this dataset, the proposed algorithm is superior with respect to other robust estimators as well as to designated vanishing point estimation algorithms. 
For self-supervised learning of the search, we evaluate the proposed algorithm on multi-homography estimation and demonstrate an accuracy that is superior to state-of-the-art methods. 

\end{abstract}
\vspace{-1.5em}
%%%%%%%%% BODY TEXT
\section{Introduction}

\begin{figure}	
\centering
\includegraphics[width=0.9\linewidth]{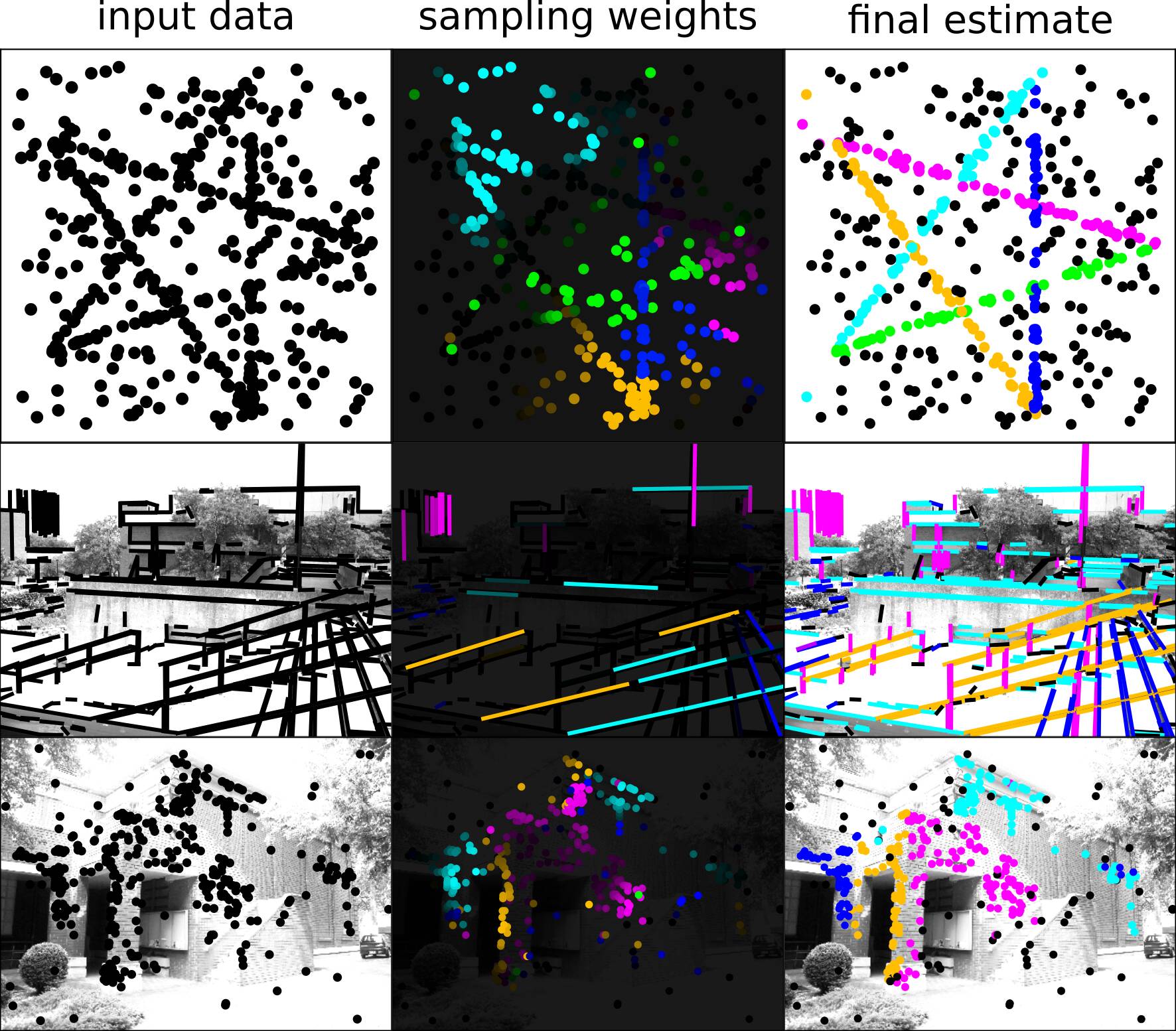}
\caption{ \textbf{CONSAC applications:} line fitting (top), vanishing point estimation (middle) and homography estimation (bottom) for multiple instances. Colour hues in column two and three indicate different instances, brightness in column two varies by sampling weight.
}
\vspace{-1.5em}
\label{fig:teaser}
\end{figure}   
Describing 3D scenes by low-dimensional parametric models, oftentimes building upon simplifying assumptions, has become fundamental to reconstructing and understanding the world around us.
Examples include: i) fitting 3D-planes to an architectural scene, which relates to finding multiple homographies in two views; ii) tracking rigid objects in two consecutive images, which relates to fitting multiple fundamental matrices; iii) identifying the dominant directions in a man-made environment, which relates to finding multiple vanishing points.  
Once such parametric models are discovered from images, they can ultimately be used in a multitude of applications and high-level vision tasks. 
Examples include the automatic creation of 3D models \cite{Agarwal2011rome,reconworld15,schoenberger2016sfm,Wandt2019RepNet}, autonomous navigation \cite{orbslam22017,sattler2016efficient,garcia2015high, kluger2020temporally} or augmented reality \cite{brachmann2016,brachmann2018lessmore,alhaija2018augmented,rameau2016real}.

Model-fitting has generally been realised as a two-step procedure. Firstly, an error-prone, low-level process to extract data points which shall adhere to a model is executed.
For example, one could match 2D feature points between pairs of images as a basis for homography estimation \cite{mutliview2004}, in order to determine the 3D plane where the 3D points live on.  
Secondly, a robust estimator that fits model parameters to \emph{inlier} data points is used, while at the same time identifying erroneous data points as so-called \emph{outliers} \cite{ransac1981}.
Some outliers can be efficiently removed by pre-processing, \eg based on the descriptor distances in feature matching \cite{Lowesift}. 

While the case of fitting a \emph{single} parametric model to data has received considerable attention in the literature, we focus on the scenario of fitting \emph{multiple} models of the same form to data. 
This is of high practical relevance, as motivated in the example above. 
There, multiple 3D planes represented by multiple homographies are fitted. 
However, when multiple models are present in the data, estimation becomes more challenging.
Inliers of one model constitute outliers of all other models. 
Naturally, outlier filters fail in removing such \emph{pseudo-outliers}.

Early approaches to multi-model fitting work \emph{sequentially}: 
They apply a robust estimator like RANSAC repeatedly, removing the data points associated with the currently predicted model in each iteration \cite{vincent2001seqransac}.
Modern, state-of-the-art methods solve multi-model fitting \emph{simultaneously} instead, by using clustering or optimisation techniques to assign data points to models or an outlier class~\cite{barath2018multi, barath2019progressive, barath2016multi, pham2014interacting, amayo2018geometric, isack2012energy,toldo2008robust,magri2014t,magri2015robust,magri2016multiple,magri2019fitting,chin2009robust}. 
In our work, we revisit the idea of sequential processing, but combine it with recent advances in learning robust estimators \cite{goodcorr18, deepfund18, brachmann2019neural}. 
Sequential processing easily lends itself to conditional sampling approaches, and with this we are able to achieve state-of-the-art results despite supposedly being conceptually inferior to simultaneous approaches. 

The main inspiration of our work stems from the work of Brachmann and Rother~\cite{brachmann2019neural}, where they train a neural network to enhance the sample efficiency of a RANSAC estimator for \emph{single model estimation}.
In contrast, we investigate \emph{multi-model fitting} by letting the neural network update sampling weights conditioned on models it has already found.  
This allows the neural network to not only suppress outliers, but also inliers of all but the current model of interest.
Since our new RANSAC variant samples model hypotheses based on conditional probabilities, we name it \emph{Conditional Sample Consensus} or CONSAC, in short.
CONSAC, as illustrated by Fig.~\ref{fig:teaser}, proves to be powerful and achieves top performance for several applications. 

Machine learning has been applied in the past to fitting of a single parametric model, by directly predicting model parameters from images \cite{kendall2015convolutional, detone16homographynet}, replacing a robust estimator \cite{goodcorr18, deepfund18, sun2020attentive} or enhancing a robust estimator \cite{brachmann2019neural}.
However, to the best of our knowledge, CONSAC is the first application of machine learning to robust fitting of \emph{multiple models}.

One limiting factor of applying machine learning to multi-model fitting is the lack of suitable datasets.
Previous works either evaluate on synthetic toy data~\cite{toldo2008robust} or few hand-labeled, real examples~\cite{wong2011dynamic, tron2007benchmark, denis2008efficient}.
The most comprehensive and widely used dataset, AdelaideRMF \cite{wong2011dynamic} for homography and fundamental matrix estimation, does not provide training data.
Furthermore, the test set consists of merely 38 labeled image pairs, re-used in various publications since $2011$ with the danger of steering the design of new methods towards overfitting to these few examples.

We collected a new dataset for multi-model fitting, vanishing point (VP) estimation in this case, which we call \textit{NYU-VP}\footnote{Code and datasets: \url{https://github.com/fkluger/consac}}.
Each image is annotated with up to eight vanishing points, and we provide pre-extracted line segments which act as data points for a robust estimator.
Due to its size, our dataset is the first to allow for supervised learning of a multi-model fitting task.
We observe that robust estimators which work well for AdelaideRMF \cite{wong2011dynamic}, do not necessarily achieve good results for our new dataset. 
CONSAC not only exceeds the accuracy of these alternative robust estimators for vanishing point estimation.
It also surpasses designated vanishing point estimation algorithms, which have access to the full RGB image instead of only pre-extracted line segments, on two datasets.

Furthermore, we demonstrate that CONSAC can be trained self-supervised for the task of multi-homography estimation, \ie where no ground truth labelling is available.
This allows us to compare CONSAC to previous robust estimators on the AdelaideRMF \cite{wong2011dynamic} dataset despite the lack of training data.
Here, we also achieve a new state-of-the-art in terms of accuracy.

\noindent To summarise, our \textbf{main contributions} are as follows:
\begin{compactitem}
\item CONSAC, the first learning-based method for robust \emph{multi-model fitting}. It is based on a neural network that sequentially updates the conditional sampling probabilities for the hypothesis selection process.
\item A new dataset, which we term \textit{NYU-VP}, for vanishing point estimation. It is the first dataset to provide sufficient training data for supervised learning of a multi-model fitting task. In addition, we present \emph{YUD+}, an extension to the York Urban Dataset \cite{denis2008efficient} (YUD) with extra vanishing point labels.
\item We achieve state-of-the-art results for vanishing point estimation for our new NYU-VP and YUD+ datasets. We exceed the accuracy of competing robust estimators as well as designated VP estimation algorithms.
\item We achieve state-of-the-art results for multi-model homography estimation on the AdelaideRMF \cite{wong2011dynamic} dataset, while training CONSAC self-supervised with an external corpus of data.
\end{compactitem}

\section{Related Work}
\subsection{Multi-Model Fitting}
Robust model fitting is a key problem in Computer Vision, which has been studied extensively in the past. 
RANSAC~\cite{ransac1981} is arguably the most commonly implemented approach. 
It samples minimal sets of observations to generate model hypotheses, computes the consensus sets for all hypotheses, \ie observations which are consistent with a hypothesis and thus inliers, and selects the hypothesis with the largest consensus. 
While effective in the single-instance case, RANSAC cannot estimate multiple model instances apparent in the data. 
Sequential RANSAC~\cite{vincent2001seqransac} fits multiple models sequentially by applying RANSAC, removing inliers of the selected hypothesis, and repeating until a stopping criterion is reached. 
PEARL~\cite{isack2012energy} instead fits multiple models simultaneously by optimising an energy-based functional, initialised via a stochastic sampling such as RANSAC. 
Several approaches based on fundamentally the same paradigm have been proposed subsequently~\cite{barath2018multi, barath2019progressive, barath2016multi, pham2014interacting, amayo2018geometric}. 
Multi-X~\cite{barath2018multi} is a generalisation to multi-class problems -- \ie cases where models of multiple types may fit the data -- with improved efficiency, while Progressive-X~\cite{barath2019progressive} interleaves sampling and optimisation in order to guide hypothesis generation using intermediate estimates. 
Another group of methods utilises preference analysis~\cite{zhang2006nonparametric} which assumes that observations explainable by the same model instance have similar distributions of residuals \wrt model hypotheses~\cite{toldo2008robust,magri2014t,magri2015robust,magri2016multiple,magri2019fitting,chin2009robust}. 
T-Linkage~\cite{magri2014t} clusters observations by their preference sets agglomeratively, with MCT~\cite{magri2019fitting} being its multi-class generalisation, while RPA~\cite{magri2015robust} uses spectral clustering instead. 
In order to better deal with intersecting models, RansaCov~\cite{magri2016multiple} formulates multi-model fitting as a set coverage problem.
Common to all of these multi-model fitting approaches is that they mostly focus on the analysis and selection of sampled hypotheses, with little attention to the sampling process itself. 
Several works propose improved sampling schemes to increase the likelihood of generating accurate model hypotheses from all-inlier minimal sets~\cite{brachmann2019neural, barath2018graph, nasuto2002napsac, chum2005matching, torr2000mlesac} in the single-instance case.
Notably, Brachmann and Rother \cite{brachmann2019neural} train a neural network to enhance the sample efficiency of RANSAC by assigning sampling weights to each data point, effectively suppressing outliers. 
Few works, such as the conditional sampling based on residual sorting by Chin et al.~\cite{chin2011accelerated}, or the guided hyperedge sampling of Purkait et al.~\cite{purkait2014clustering}, consider the case of multiple instances. 
In contrast to these hand-crafted methods, we present the first learning-based conditional sampling approach.

\subsection{Vanishing Point Estimation}
While vanishing point (VP) estimation is part of a broader spectrum of multi-model fitting problems, a variety of algorithms specifically designed to tackle this task has emerged in the past~\cite{antunes2013global, barinova2010geometric, kluger2017deep, lezama2014finding, simon2018acontrario, tardif2009non, vedaldi2012self, wildenauer2012robust, xu2013minimum, zhai2016detecting}. 
While most approaches proceed similarly to other multi-model fitting methods, they usually exploit additional, domain-specific knowledge. Zhai et al.~\cite{zhai2016detecting} condition VP estimates on a horizon line, which they predict from the RGB image via a convolutional neural network (CNN). 
Kluger et al.~\cite{kluger2017deep} employ a CNN which predicts initial VP estimates, and refine them using a task-specific expectation maximisation~\cite{dempster1977maximum} algorithm. Simon et al.~\cite{simon2018acontrario} condition the VPs on the horizon line as well. 
General purpose robust fitting methods, such as CONSAC, do not rely on such domain-specific constraints.
Incidentally, these works on VP estimation conduct evaluation using a metric which is based on the horizon line instead of the VPs themselves. 
As there can only be one horizon line per scene, this simplifies evaluation in presence of ambiguities \wrt the number of VPs, but ultimately conceals differences in performance regarding the task these methods have been designed for. 
By comparison, we conduct evaluation on the VPs themselves.

\section{Method}
\begin{figure}	
\centering
\includegraphics[width=0.95\linewidth]{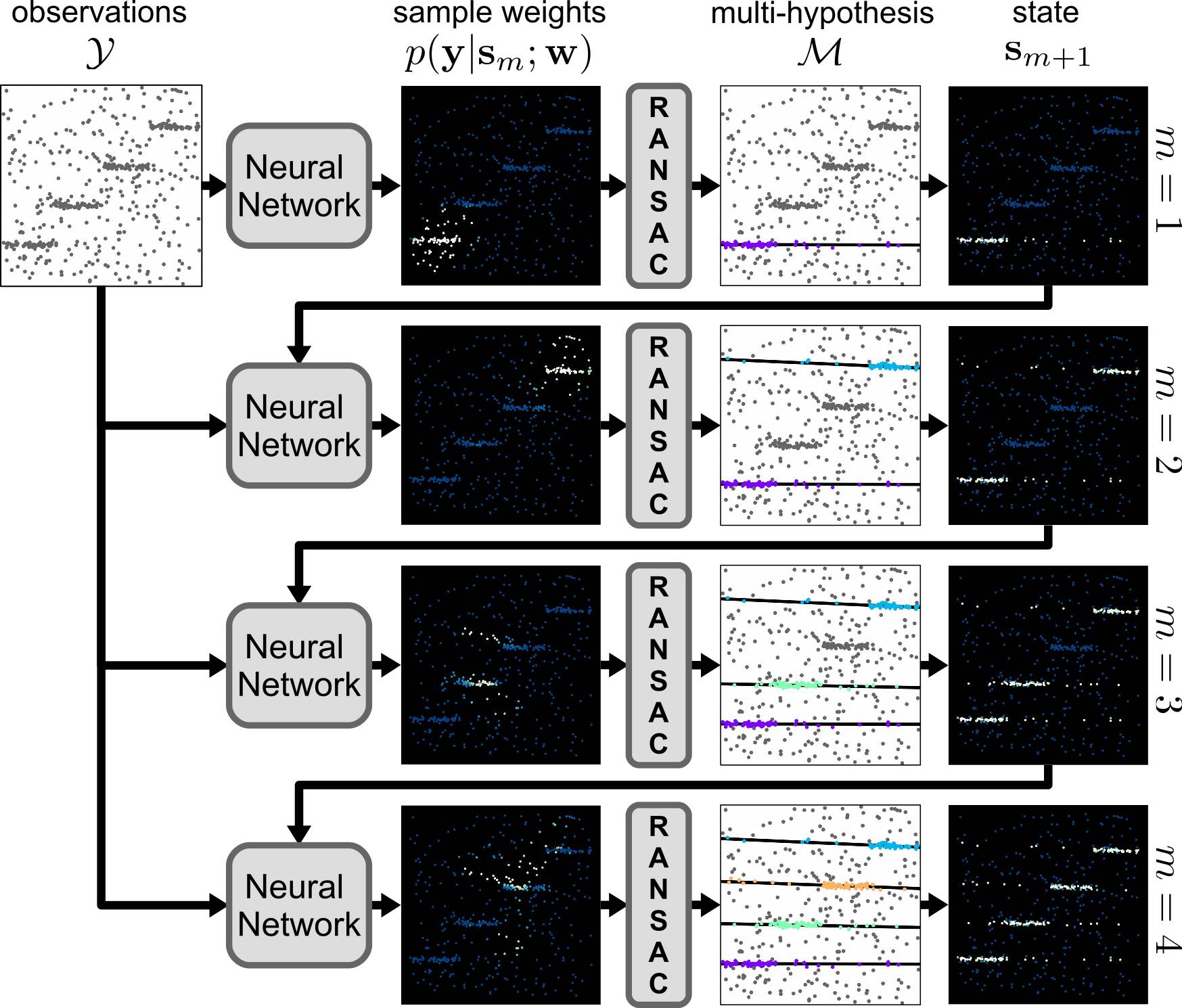}
\caption{ \textbf{Multi-Hypothesis Generation:} a neural network predicts sampling weights $p$ for all observations conditioned on a state $\vec{s}$. A RANSAC-like sampling process uses these weights to select a model hypothesis and appends it to the current multi-instance hypothesis $\set{M}$. The state $\vec{s}$ is updated based on $\set{M}$ and fed into the neural network repeatedly.}
\label{fig:system}
\vspace{-1em}
\end{figure}   
Given a set of noisy observations $\vec{y} \in \set{Y}$ contaminated by outliers, we seek to fit $M$ instances of a geometric model $\vec{h}$ apparent in the data. 
We denote the set of all model instances as  $\set{M} = \{\vec{h}_1, \dots, \vec{h}_M \}$.
CONSAC estimates $\set{M}$ via three nested loops, cf. Fig.~\ref{fig:system}. 

\begin{compactenum}
\item We generate a single model instance $\hat{\vec{h}}$ via RANSAC-based~\cite{ransac1981} sampling, guided by a neural network.
This level corresponds to one row of Fig.~\ref{fig:system}.
\item We repeat single model instance generation while conditionally updating sampling weights.
Multiple single model hypotheses compound to a \emph{multi-hypothesis} $\set{M}$. 
This level corresponds to the entirety of Fig.~\ref{fig:system}.
\item We repeat steps 1 and 2 to sample multiple multi-hypotheses $\set{M}$ independently. 
We choose the best multi-hypothesis as the final multi-model estimate $\hat{\set{M}}$.
\end{compactenum}
\noindent We discuss these conceptional levels more formally below.
\vspace{-.5em}
\paragraph{Single Model Instance Sampling}
We estimate parameters of a single model, \eg one VP, from a minimal set of $C$ observations, \eg two line segments, using a minimal solver $f_{\mathsf{S}}$.
As in RANSAC, we compute a hypothesis pool  $\set{H} = \{ \vec{h}_1, \dots, \vec{h}_S \}$ via random sampling of $S$ minimal sets. 
We choose the best hypothesis $\hat{\vec{h}}$ based on a \emph{single-instance} scoring function $g_{\mathsf{s}}$. 
Typically, $g_{\mathsf{s}}$ is realised as inlier counting via a residual function $r(\vec{y}, \vec{h})$ and a threshold $\tau$. 
\vspace{-.5em}
\paragraph{Multi-Hypothesis Generation}
We repeat single model instance sampling $M$ times to generate a full multi-hypothesis $\set{M}$, \eg a complete set of vanishing points for an image.
Particularly, we select $M$ model instances $\hat{\vec{h}}_m$ from their respective hypothesis pools $\set{H}_m$.
Applied sequentially, previously chosen hypotheses can be factored into the scoring function $g_{\mathsf{s}}$ when selecting $\hat{\vec{h}}_m$:
\begin{equation}
\hat{\vec{h}}_m = \argmax_{\vec{h} \in \set{H}_m} g_{\mathsf{s}}(\vec{h}, \set{Y}, \hat{\vec{h}}_{1:(m-1)}) \, .
\end{equation}

\paragraph{Multi-Hypothesis Sampling}
We repeat the previous process $P$ times to generate a pool of multi-hypotheses $\set{P} = \{ \set{M}_1, \dots \set{M}_P \}$.
We select the best multi-hypothesis according to a multi-instance scoring function $g_{\mathsf{m}}$:
\begin{equation}
\hat{\set{M}} = \argmax_{\set{M} \in \set{P}} g_{\mathsf{m}}(\set{M}, \set{Y}) \, ,
\end{equation}
where $g_{\mathsf{m}}$ measures the joint inlier count of all hypotheses in $\set{M}$, and where the $\mathsf{m}$ in $g_{\mathsf{m}}$ stands for \emph{multi-instance}.

\subsection{Conditional Sampling}
RANSAC samples minimal sets uniformly from $\set{Y}$. 
For large amounts of outliers in $\set{Y}$, the number of samples $S$ required to sample an outlier-free minimal set with reasonable probability grows exponentially large. 
Brachmann and Rother~\cite{brachmann2019neural} instead sample observations according to a categorical distribution $\vec{y} \sim p(\vec{y}; \vec{w})$ parametrised by a neural network $\vec{w}$.
The neural network biases sampling towards outlier-free minimal sets which generate accurate hypotheses $\vec{\hat{h}}$. 
While this approach is effective in the presence of outliers, it is not suitable for dealing with pseudo-outliers posed by multiple model instances. 
Sequential RANSAC~\cite{vincent2001seqransac} conditions the sampling on previously selected hypotheses, \ie $\vec{y} \sim p(\vec{y} | \{ \vec{\hat{h}}_1, \dots, \vec{\hat{h}}_{m-1} \})$, by removing observations already deemed as inliers from $\set{Y}$ after each hypothesis selection. 
While being able to reduce pseudo-outliers for subsequent instances, this approach can neither deal with pseudo-outliers in the first sampling step, nor with gross outliers in general. 
Instead, we parametrise the conditional distribution by a neural network $\vec{w}$ conditioned on a state $\vec{s}$: $\vec{y} \sim p(\vec{y} | \vec{s}; \vec{w}) \, .$

The state vector $\vec{s}_m$ at instance  sampling step $m$ encodes information about previously sampled hypotheses in a meaningful way. 
We use the inlier scores of all observations \wrt all previously selected hypotheses as the state $\vec{s}_m$.
We define the state entry $s_{m,i}$ of observation $\vec{y}_i$ as:
\begin{equation}
    s_{m,i} = \max_{j \in [1,m)} g_{\mathsf{y}}(\vec{y}_i, \hat{\vec{h}}_j) \, ,
    \label{eq:inlier_score}
\end{equation}
with $g_{\mathsf{y}}$ gauging if $\vec{y}$ is an inlier of model $\vec{h}$. 
See the last column of Fig.~\ref{fig:system} for a visualisation of the state.
We sample multi-instance hypothesis pools independently:
\begin{equation}
p(\set{P}; \vec{w}) = \prod_{i=1}^P p(\set{M}_i; \vec{w}) \, ,
\end{equation}
while conditioning multi-hypotheses on the state $\vec{s}$:
\begin{equation}
\begin{split}
 p(\set{M}; \vec{w}) & = \prod_{m=1}^M p({\set{H}}_m | \vec{s}_m; \vec{w}) \, , \\
\text{with} \quad  p(\set{H} | \vec{s}; \vec{w}) & = \prod_{s=1}^S p({\vec{h}}_s | \vec{s}; \vec{w}) \, , \\
\text{with} \quad  p(\vec{h} | \vec{s}; \vec{w}) & = \prod_{c=1}^C p(\vec{y}_c | \vec{s}; \vec{w}) \, .
\end{split}
\end{equation}
Note that we do not update state $\vec{s}$ while sampling single instance hypotheses pools $\set{H}$, but only within sampling of multi-hypotheses $\set{M}$.
We provide details of scoring functions $g_{\mathsf{y}}$, $g_{\mathsf{m}}$ and $g_{\mathsf{s}}$ in the appendix.

\subsection{Neural Network Training}
Neural network parameters $\vec{w}$ shall be optimised in order to increase chances of sampling outlier- and pseudo-outlier-free minimal sets which result in accurate, complete and duplicate-free multi-instance estimates $\set{\hat{M}}$. 
As in \cite{brachmann2019neural}, we minimise the expectation of a task loss $\ell ({\set{\hat{M}}})$ which measures the quality of an estimate:
\begin{equation}
    \mathcal{L}(\vec{w}) = \mathbb{E}_{\set{P} \sim p(\set{P}; \vec{w})} \left[ \ell ({\set{\hat{M}}}) \right] \, .
    \label{eq:exp_task_loss}
\end{equation}
In order to update the network parameters $\vec{w}$, we approximate the gradients of the expected task loss:
\begin{equation}
    \frac{\partial}{\partial \vec{w}}  \mathcal{L}(\vec{w}) = \mathbb{E}_{\set{P}} \left[ \ell ({\set{\hat{M}}}) \frac{\partial}{\partial \vec{w}} \log p(\set{P}; \vec{w}) \right] \, ,
    \label{eq:exp_task_loss_grad}
\end{equation}
by drawing $K$ samples $\set{P}_k \sim p(\set{M}; \vec{w})$:
\begin{equation}
    \frac{\partial}{\partial \vec{w}}  \mathcal{L}(\vec{w}) \approx \frac{1}{K} \sum_{k=1}^K \left[ \ell ({\set{\hat{M}}_k}) \frac{\partial}{\partial \vec{w}} \log p(\set{P}_k; \vec{w}) \right] \, .
    \label{eq:exp_task_loss_grad_approx}
\end{equation}
As we can infer from Eq.~\ref{eq:exp_task_loss_grad}, neither the loss $\ell$,  nor the sampling procedure for $\set{\hat{M}}$ need be differentiable. 
As in~\cite{brachmann2019neural}, we subtract the mean loss from $\ell$ to reduce variance.

\subsubsection{Supervised Training}

If ground truth models $\set{M}^{\mathsf{gt}} = \{\vec{h}^{\mathsf{gt}}_1, \dots, \vec{h}^{\mathsf{gt}}_G\}$ are available, we can utilise a task-specific loss $\ell_{\mathsf{s}}(\vec{\hat{h}}, \vec{h}^{\mathsf{gt}})$ measuring the error between a single ground truth model $\vec{m}$ and an estimate $\vec{\hat{h}}$. 
For example, $\ell_{\mathsf{s}}$ may measure the angle between an estimated and a true vanishing direction. 
First, however, we need to find an assigment between $\set{M}^{\mathsf{gt}}$ and $\set{\hat{M}}$. 
We compute a cost matrix $\vec{C}$, with $C_{ij} = \ell_{\mathsf{s}}(\hat{\vec{h}}_i, \vec{h}^{\mathsf{gt}}_j) \,$,
and define the multi-instance loss as the minimal cost of an assignment obtained via the Hungarian method~\cite{kuhn1955hungarian} $f_{\mathsf{H}}$: $\ell(\hat{\set{M}}, \set{M}^{\mathsf{gt}}) = f_{\mathsf{H}}(\vec{C}_{1:\min(M,G)}) \, .$
Note that we only consider \emph{at most} $G$ model estimates $\vec{\hat{h}}$ which have been selected first, regardless of how many estimates $M$ were generated, \ie this loss encourages early selection of good model hypotheses, but does not penalise bad hypotheses later on.

\subsubsection{Self-supervised Training}
\label{subsubsec:self_supervised}
In absence of ground-truth labels, we can train CONSAC in a self-supervised fashion by replacing the task loss with another quality measure. 
We aim to maximise the average joint inlier counts of the selected model hypotheses:
\begin{equation}
    g_{\mathsf{ci}}(\hat{\vec{h}}_m, \set{Y}) = \frac{1}{|\set{Y}|} \sum_{i=1}^{|\set{Y}|} \max_{j \in [1,m]} g_{\mathsf{i}}(\vec{y}_i, \hat{\vec{h}}_j) \, .
    \label{eq:avg_cumul_inlier_count}
\end{equation}
We then define our self-supervised loss as:
\begin{equation}
    \ell_{\mathsf{self}}(\hat{\set{M}}) = - \frac{1}{M} \sum_{m=1}^M  g_{\mathsf{ci}}(\hat{\vec{h}}_m, \set{Y}) \, .
\end{equation}
Eq.~\ref{eq:avg_cumul_inlier_count} monotonically increases \wrt $m$, and has its minimum when the models in $\set{\hat{M}}$ induce the largest possible minimally overlapping inlier sets descending in size. 

\paragraph{Inlier Masking Regularisation}
For self-supervised training, we found it empirically beneficial to add a weighted regularisation term $\kappa \cdot \ell_{\mathsf{im}}$ penalising large sampling weights for observations $\vec{y}$ which have already been recognised as inliers: $\ell_{\mathsf{im}}(\tilde{p}_{m,i}) = \max(0,\, \tilde{p}_{m,i} + s_{m,i} - 1) \, ,$ 
with $s_{m,i}$ being the inlier score as per Eq.~\ref{eq:inlier_score} for observation $\vec{y}_i$ at instance sampling step $m$, and $\tilde{p}_{m,i}$ being its normalised sampling weight:
\begin{equation}
    \tilde{p}_{m,i} =  \frac{p(\vec{y}_i | \vec{s}_m; \vec{w})}{\max_{{\vec{y}} \in \set{Y}} p({\vec{y}} | \vec{s}_m; \vec{w})} \, .
\end{equation}

\subsection{Post-Processing at Test Time}
\label{subsec:refinement}
\paragraph{Expectation Maximisation}
In order refine the selected model parameters $\set{\hat{M}}$, we implement a simple EM~\cite{dempster1977maximum} algorithm. 
Given the posterior distribution:
\begin{equation}
    p(\vec{h} | \vec{y}) = \frac{p(\vec{y} | \vec{h}) p(\vec{h})}{p(\vec{y})}, \, \text{with} \,\, p(\vec{y}) = \sum_{m=1}^M p(\vec{y} | \vec{h}_m)\, ,
\end{equation}
and likelihood $ p(\vec{y} | \vec{h}) = \sigma\inverse \phi(r(\vec{y}, \vec{h}) \sigma\inverse)$ modelled by a normal distribution,
we optimise model parameters $\set{M}^*$ such that $\set{M}^* = \argmax_{\set{M}} p(\set{Y})$ with:
\begin{equation}
 p(\set{Y}) = \prod_{i=1}^{|\set{Y}|} \sum_{m=1}^{M} p(\vec{y}_i | \vec{h}_m)p(\vec{h}_m) \, ,
\end{equation}
using fixed $\sigma$ and $p(\vec{h})=1$ for all $\vec{h}$.

\paragraph{Instance Ranking}
In order to asses the significance of each selected model instance $\vec{\hat{h}}$, we compute a permutation $\pmb{\pi}$ greedily sorting $\set{\hat{M}}$ by joint inlier count, \ie:
\begin{equation}
    \pi_m = \argmax_q \sum_{i=1}^{|\set{Y}|} \max_{j \in \pmb{\pi}_{1:m-1} \cup \{ q\}} g_{\mathsf{i}}(\vec{y}_i, \hat{\vec{h}}_j) \, .
\end{equation}
Such an ordering is useful in applications where the true number of instances present in the data may be ambiguous, and less significant instances may or may not be of interest. 
Small objects in a scene, for example, may elicit their own vanishing points, which may appear spurious for some applications, but could be of interest for others.

\paragraph{Instance Selection}
In some scenarios, the number of instances $M$ needs to be determined as well but is not known beforehand, \eg for uniquely assigning observations to model instances.
For such cases, we consider the subset of instances $\hat{\set{M}}_{1:q}$ up to the $q$-th model instance $\hat{\vec{h}}_q$ which increases the joint inlier count by at least $\Theta$. 
Note that the inlier threshold $\theta$ for calculating the joint inlier count at this point may be chosen differently from the inlier threshold $\tau$ during hypothesis sampling.
For example, in our experiments for homography estimation, we use a $\theta > \tau$ in order to strike a balance between under- and oversegmentation.

\section{Multi-Model Fitting Datasets}
Robust multi-model fitting algorithms can be applied to various tasks. 
While earlier works mostly focused on synthetic problems, such as fitting lines to point sets artificially perturbed by noise and outliers~\cite{toldo2008robust}, real-world datasets for other tasks have been used since. 
The AdelaideRMF~\cite{wong2011dynamic} dataset contains $38$ image pairs with pre-computed SIFT~\cite{Lowesift} feature point correspondences, which are clustered either via homographies (same plane) or fundamental matrices (same motion). 
Hopkins155~\cite{tron2007benchmark} consists of 155 image sequences with on average 30 frames each. 
Feature point correspondences are given as well, also clustered via their respective motions. 
For vanishing point estimation, the York Urban Dataset (YUD)~\cite{denis2008efficient} contains $102$ images with three orthogonal ground truth vanishing directions each. 
All these datasets have in common that they are very limited in size, with no or just a small portion of the data reserved for training or validation. 
As a result, they are easily susceptible to parameter overfitting and ill-suited for contemporary machine learning approaches. 

\begin{figure}	
		\centering
		\begin{subfigure}[t]{0.3\linewidth}
			\centering
			\includegraphics[width=0.99\linewidth]{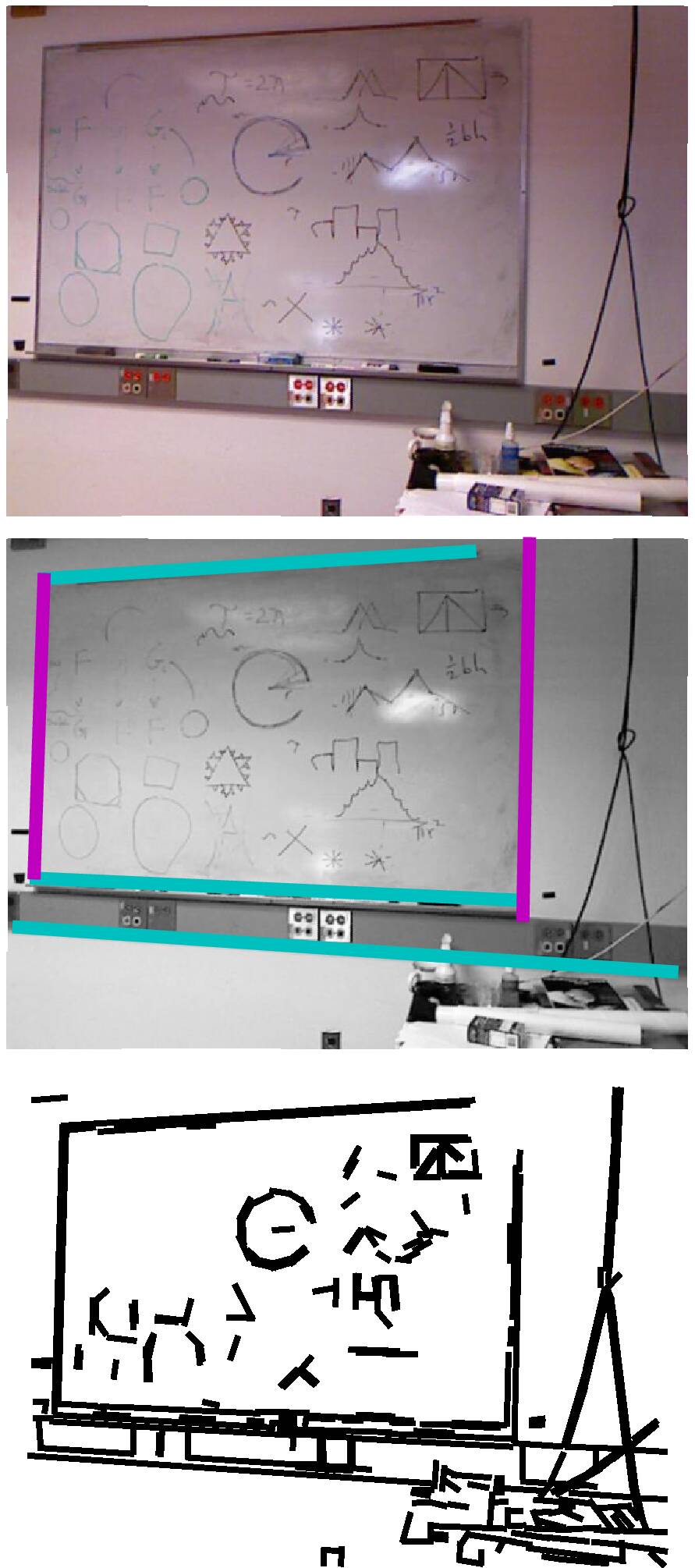}
		\end{subfigure}
		\begin{subfigure}[t]{0.3\linewidth}
			\centering
			\includegraphics[width=0.99\linewidth]{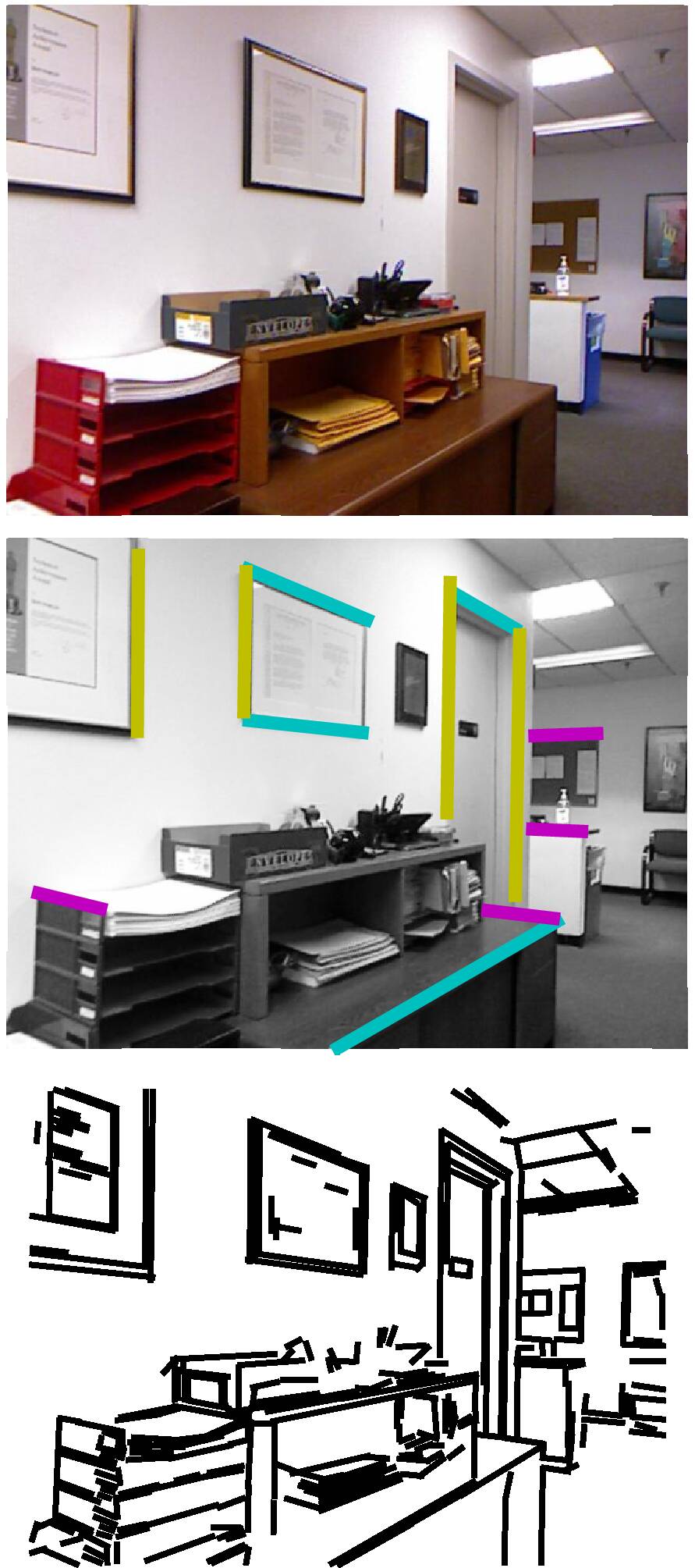}
		\end{subfigure}
		\begin{subfigure}[t]{0.3\linewidth}
			\centering
			\includegraphics[width=0.99\linewidth]{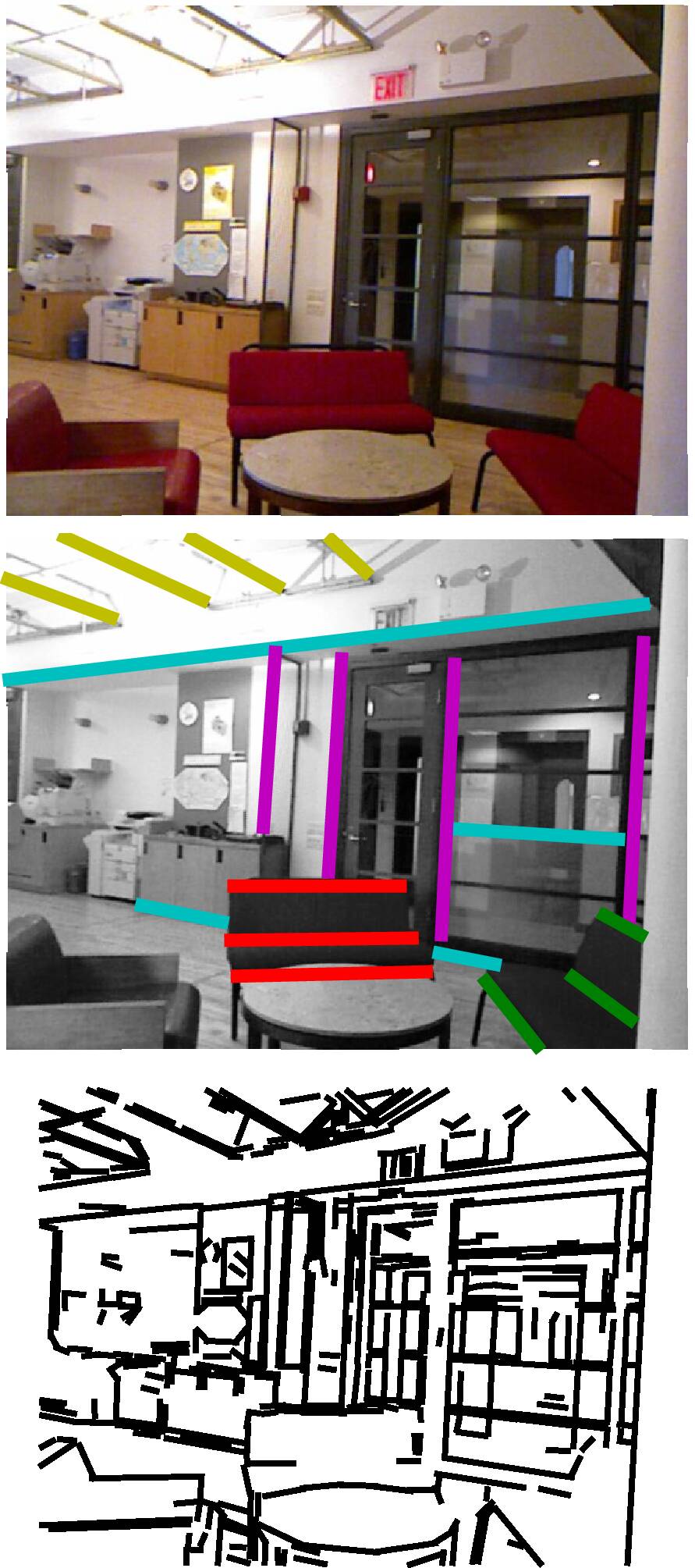}
		\end{subfigure}
		\caption{Examples from our newly presented NYU-VP dataset with two (left), three (middle) and five (right) vanishing points. \textbf{Top:} Original RGB image. \textbf{Middle:} Manually labelled line segments used to generate ground truth VPs. \textbf{Bottom:} Automatically extracted line segments. }
		\label{fig:nyu_examples}
\vspace{-.8em}
\end{figure}

\paragraph{NYU Vanishing Point Dataset}
We therefore introduce the \textit{NYU-VP} dataset. 
Based on the NYU Depth V2~\cite{silberman2012indoor} (NYU-D) dataset, it contains ground truth vanishing point labels for $1449$ indoor scenes, \ie it is more than ten times larger than the previously largest dataset in its category; see Tab.~\ref{tab:dataset_comparison} for a comparison. To obtain each VP, we manually annotated at least two corresponding line segments. 
While most scenes show three VPs, it ranges between one and eight. 
In addition, we provide line segments extracted from the images with LSD~\cite{von2008lsd}, which we used in our experiments. 
Examples are shown in Fig.~\ref{fig:nyu_examples}. 

\paragraph{YUD+} 
Each scene of the original York Urban Dataset (YUD)~\cite{denis2008efficient} is labelled with exactly three VPs corresponding to orthogonal directions consistent with the Manhattan-world assumption. 
Almost a third of all scenes, however, contain up to five additional significant yet unlabelled VPs. 
We labelled these VPs in order to allow for a better evaluation of VP estimators which do not restrict themselves to Manhattan-world scenes. 
This extended dataset, which we call \emph{YUD+}, will be made available together with the automatically extracted line segments used in our experiments.

\begin{table}
    \centering
    \begin{tabular}{c|c|c|c|c}
        \small{task} & \small{dataset} & \small{train+val} & \small{test} & \small{instances} \\
        \hline
        H                       & Adelaide~\cite{wong2011dynamic}    & $0$   & $19$  & 1--6 \\
        \hline
        \multirow{2}{*}{F}     & Adelaide~\cite{wong2011dynamic}   & $0$   & $19$  & 1--4 \\
                               & Hopkins~\cite{tron2007benchmark}  & $0$   & $155$ & 2--3 \\
       \hline
        \multirow{3}{*}{VP}     & YUD~\cite{denis2008efficient}         & $25$  & $77$  & 3 \\
                                & YUD+ (ours)                           & $25$  & $77$  & 3--8 \\
                                & NYU-VP (ours)                         & $1224$& $225$ & 1--8 \\
                                       
    \end{tabular}
    \caption{Comparison of datasets for different applications of multi-model fitting: vanishing point (VP), homography (H) and fundamental matrix (F) fitting. We compare the numbers of combined training and validation scenes, test scenes, and model instances per scene.}
    \label{tab:dataset_comparison}
    \vspace{-1em}
\end{table}

\section{Experiments}
For conditional sampling weight prediction, we implement a neural network based on the architecture of \cite{brachmann2019neural, goodcorr18}. 
We provide implementation and training details, as well as more detailed experimental results, in the appendix.

\subsection{Line Fitting}
\begin{figure}[b]
\centering
\includegraphics[width=0.99\linewidth]{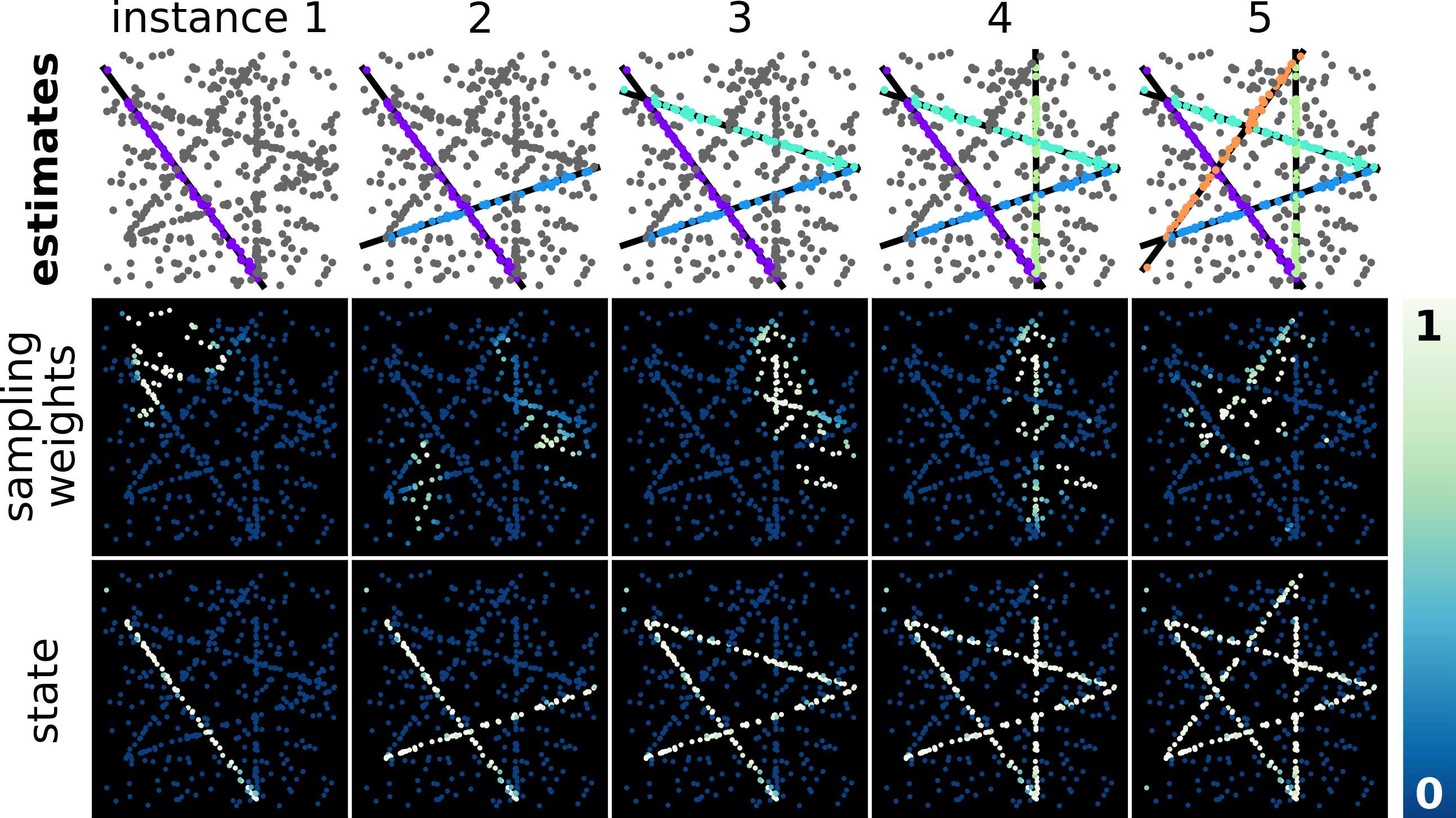}
\caption{\textbf{Line fitting} result for the \emph{star5} scene from \cite{toldo2008robust}. We show the generation of the multi-hypothesis $\hat{\set{M}}$ eventually selected by CONSAC. \textbf{Top:} Original points with estimated line instances at each instance selection step. \textbf{Middle:} Sampling weights  at each instance step. \textbf{Bottom:} State $\vec{s}$ generated from the selected model instances.  }
\label{fig:star5}
\end{figure}

We apply CONSAC to the task of fitting multiple lines to a set of noisy points with outliers. 
For training, we generated a synthetic dataset: each scene consists of randomly placed lines with points uniformly sampled along them and perturbed by Gaussian noise, and uniformly sampled outliers.
After training CONSAC on this dataset in a supervised fashion, we applied it to the synthetic dataset of~\cite{toldo2008robust}. Fig.~\ref{fig:star5} shows how CONSAC sequentially focuses on different parts of the scene, depending on which model hypotheses have already been chosen, in order to increase the likelihood of sampling outlier-free non-redundant hypotheses.
Notably, the network learns to focus on junctions rather than individual lines for selecting the first instances.
The RANSAC-based single-instance hypothesis sampling makes sure that CONSAC still selects an individual line.

\subsection{Vanishing Point Estimation}
A vanishing point $\vec{v} \propto \vec{K}\vec{d}$ arises as the projection of a direction vector $\vec{d}$ in 3D onto an image plane using camera parameters $\vec{K}$. 
Parallel lines, \ie with the same direction $\vec{d}$, hence converge in $\vec{v}$ after projection. 
If $\vec{v}$ is known, the corresponding direction $\vec{d}$ can be inferred via inversion: $\vec{d} \propto \vec{K}\inverse \vec{v}$. 
VPs therefore provide information about the 3D structure of a scene from a single image. 
While two corresponding lines are sufficient to estimate a VP, real-world scenes generally contain multiple VP instances. 
We apply CONSAC to the task of VP detection and evaluate it on our new NYU-VP and YUD+ datasets, as well as on YUD~\cite{denis2008efficient}. 
We compare against several other robust estimators, and also against task-specific state-of-the art VP detectors. 
We train CONSAC on the training set of NYU-VP in a supervised fashion and evaluate on the test sets of NYU-VP, YUD+ and YUD using the same parameters. 
YUD and YUD+ were neither used for training nor parameter tuning. Notably, NYU-VP only depicts indoor scenes, while YUD also contains outdoor scenes.

\subsubsection{Evaluation Protocol}
We compute the error $e(\vec{\hat{h}}, \vec{h}^\mathsf{gt})$ between two particular VP instances via the angle between their corresponding directions in 3D. 
Let $\vec{C}$ be the cost matrix with $C_{ij} = e( \vec{\hat{h}}_i,\vec{h}^\mathsf{gt}_j)$. 
We can find a matching between ground truth $\set{M}^\mathsf{gt}$ and estimates $\set{\hat{M}}$ by applying the Hungarian method on $\vec{C}$ and consider the errors of the matched VP pairs. 
For $N > M$ however, this would benefit methods with a tendency to oversegment, as a larger number of estimated VPs generally increases the likelihood of finding a good match to a ground truth VP. 
On the other hand, we argue that strictly penalising oversegmentation \wrt the ground truth is unreasonable, as smaller or more fine-grained structures which may have been missed during labelling may still be present in the data. 
We therefore assume that the methods also provide a permutation $\pmb{\pi}$ (cf. Sec.~\ref{subsec:refinement}) which ranks the estimated VPs by their significance, and evaluate using at most $N$ most significant estimates. 
After matching, we generate the recall curve for all VPs of the test set and calculate the area under the curve (AUC) up to an error of $10^{\circ}$. 
We report the average AUC and its standard deviation over five runs. 

\subsubsection{Robust Estimators}
\begin{figure}	
\centering
\includegraphics[width=0.99\linewidth]{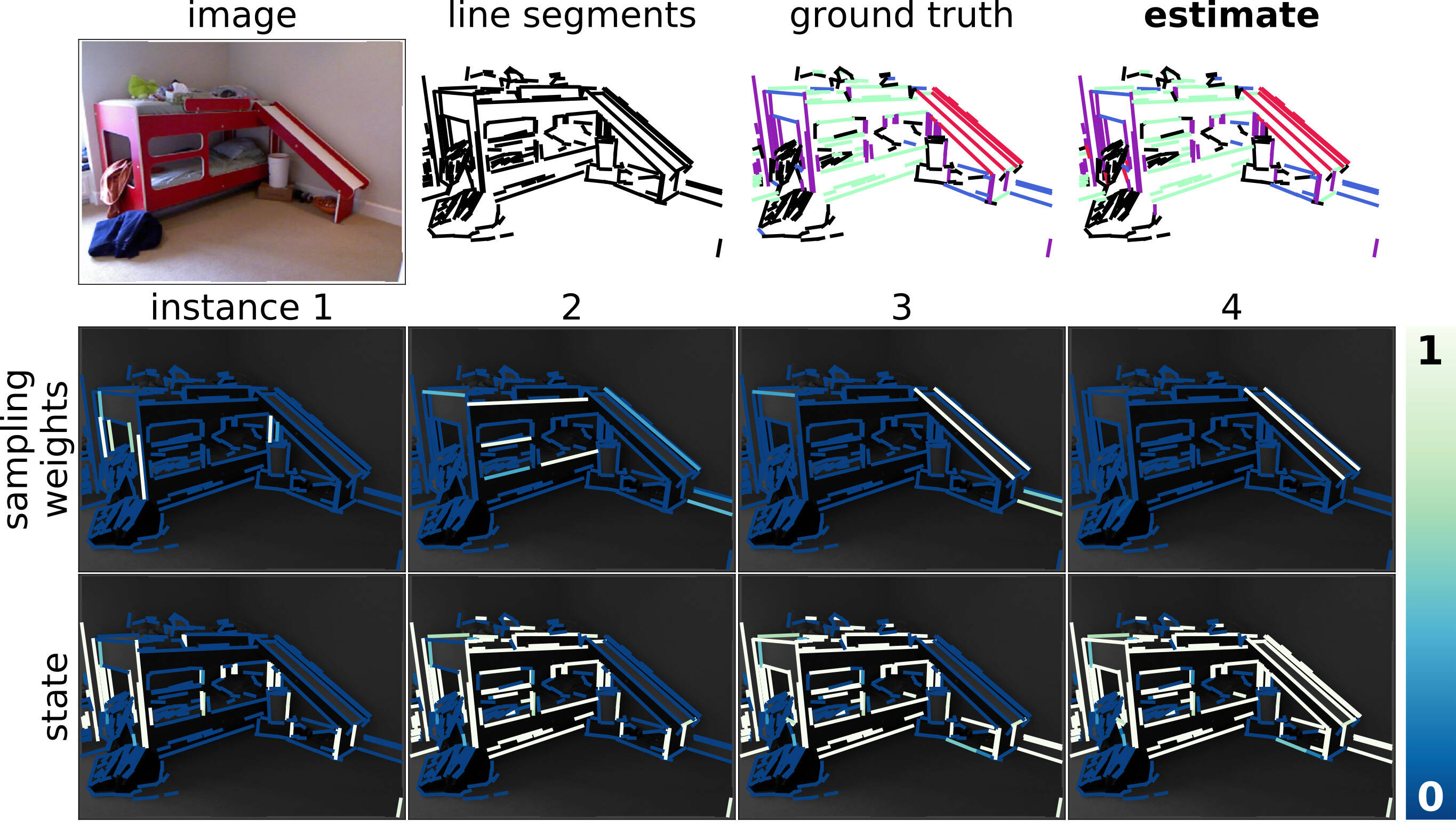}
\caption{\textbf{VP fitting} result for a scene from the NYU-VP test set. \textbf{Top:} Original image, extracted line segments, assignment to ground truth VPs, and assignment to VPs predicted by CONSAC (average error: $2.2^{\circ}$). \textbf{Middle:} Sampling weights of line segments at each instance step. \textbf{Bottom:} State $\vec{s}$ generated from the selected model instances. 
}
\label{fig:nyu_result_example}
\vspace{-.5em}
\end{figure}

We compare against T-Linkage~\cite{magri2014t}, MCT~\cite{magri2019fitting}, Multi-X~\cite{barath2018multi}, RPA~\cite{magri2015robust}, RansaCov~\cite{magri2016multiple} and Sequential RANSAC~\cite{vincent2001seqransac}. 
We used our own implementation of T-Linkage and Sequential RANSAC, while adapting the code provided by the authors to VP detection for the other methods. 
All methods including CONSAC get the same line segments (geometric information only) as input, use the same residual metric and the same inlier threshold, and obtain the permutation $\pmb{\pi}$ as described in Sec.~\ref{subsec:refinement}. 
As Tab.~\ref{tab:vp_results} shows, CONSAC outperforms its competitors on all three datasets by a large margin.
Although CONSAC was only trained on indoor scenes (NYU-VP) it also performs well on outdoor scenes (YUD/YUD+).
Perhaps surprisingly, Sequential RANSAC also performs favourably, thus defying the commonly held notion that this greedy approach does not work well. 
Fig.~\ref{fig:nyu_result_example} shows a qualitative result for CONSAC.

\subsubsection{Task-Specific Methods}
In addition to general-purpose robust estimators, we evaluate the state-of-the-art task-specific VP detectors of Zhai et al.~\cite{zhai2016detecting}, Kluger et al.~\cite{kluger2017deep} and Simon et al.~\cite{simon2018acontrario}. 
Unlike the robust estimators, these methods may use additional information, such as the original RGB image, or enforce additional geometrical constraints. 
The method of Kluger et al. provides a score for each VP, which we used to generate the permutation $\pmb{\pi}$. 
For Zhai et al. and Simon et al., we resorted to the more lenient na\"ive evaluation metric instead. 
Despite this, CONSAC performs superior to all task-specific methods on NYU-VP and YUD+, and slightly worse on YUD.

\begin{table}
\setlength\tabcolsep{.42em}
	\begin{center}
	\begin{tabular}{|l|cc|cc|cc|}
	\cline{2-7}
	\multicolumn{1}{c|}{} & \multicolumn{2}{c|}{NYU-VP}    & \multicolumn{2}{c|}{YUD+}     & \multicolumn{2}{c|}{YUD~\cite{denis2008efficient}} \\
	\cline{2-7}
	\multicolumn{1}{c|}{} &   \small{avg.} & \small{std.} & \small{avg.} & \small{std.}  & \small{avg.} & \small{std.} \\
	\hline
	\multicolumn{7}{|c|}{\small{robust estimators (on pre-extracted line segments)}} \\
	\textbf{CONSAC }                        & $\mathbf{65.0}$ & $0.46$ & $\mathbf{77.1}$ & $0.24$ & ${83.9}$ & $0.24$  \\
	T-Linkage~\cite{magri2014t}                 & $57.8$ & $0.07$ & $72.6$ & $0.67$ & $79.2$ & $0.93$ \\
	Seq. RANSAC                                 & $53.6$ & $0.40$ & $69.1$ & $0.57$ & $76.2$ & $0.75$\\
	MCT~\cite{magri2019fitting}                 & $47.0$ & $0.67$ & $62.7$ & $1.28$ & $67.7$ & $0.59$  \\
	Multi-X~\cite{barath2018multi}              & $41.3$ & $1.00$ & $50.6$ & $0.80$ & $55.3$ & $1.00$\\
	RPA~\cite{magri2015robust}                  & $39.4$ & $0.65$ & $48.5$ & $1.14$ & $52.5$ & $1.35$ \\
	RansaCov~\cite{magri2016multiple}           & $7.9$  & $0.62$ & $13.4$ & $1.76$ & $13.9$ & $1.49$\\
	\hline
	\multicolumn{7}{|c|}{\small{task-specific methods (full information)}} \\
	Zhai~\cite{zhai2016detecting}\tssdagger    & $63.0$ & $0.25$ & $72.1$ & $0.50$ & $84.2$ & $0.69$ \\
	Simon~\cite{simon2018acontrario}\tssdagger & $62.1$ & $0.67$ & $73.6$ & $0.77$ & $85.1$ & $0.74$  \\
	Kluger~\cite{kluger2017deep}               & $61.7$ & ---\tssstar& $74.7$ & ---\tssstar &  $\mathbf{85.9}$ & ---\tssstar \\
	\hline
	\end{tabular}
    \end{center}
    \vspace{-4mm}
	\caption{\textbf{VP estimation:} Average AUC values (avg., in \%, higher is better) and their standard deviations (std.) over five runs for vanishing point estimation on our new NYU-VP and YUD+ datasets as well as on YUD~\cite{denis2008efficient}. *~Not applicable. \textdagger~Na\"ive evaluation metric. }
	\label{tab:vp_results}
\vspace{-4mm}
\end{table}	

\subsection{Two-view Plane Segmentation}
Given feature point correspondences from two images showing different views of the same scene, we estimate multiple homographies $\vec{H}$ conforming to different 3D planes in the scene. 
As no sufficiently large labelled datasets exist for this task, we train our approach self-supervised (CONSAC-S) using SIFT feature correspondences extracted from the structure-from-motion scenes of~\cite{reconworld15,strecha2008benchmarking,xiao2013sun3d} also used by \cite{brachmann2019neural}. 
Evaluation is performed on the AdelaideRMF~\cite{wong2011dynamic} homography estimation dataset and adheres to the protocol used by \cite{barath2019progressive}, \ie we report the average misclassification error (ME) and its standard deviation over all scenes for five runs using identical parameters. 
We compare against the robust estimators Progressive-X~\cite{barath2019progressive}, Multi-X~\cite{barath2018multi}, PEARL~\cite{isack2012energy}, MCT~\cite{magri2019fitting}, RPA~\cite{magri2015robust}, T-Linkage~\cite{magri2014t}, RansaCov~\cite{magri2016multiple} and Sequential RANSAC~\cite{vincent2001seqransac}. 

\subsubsection{Results}
\begin{figure}	
\centering
\includegraphics[width=0.99\linewidth]{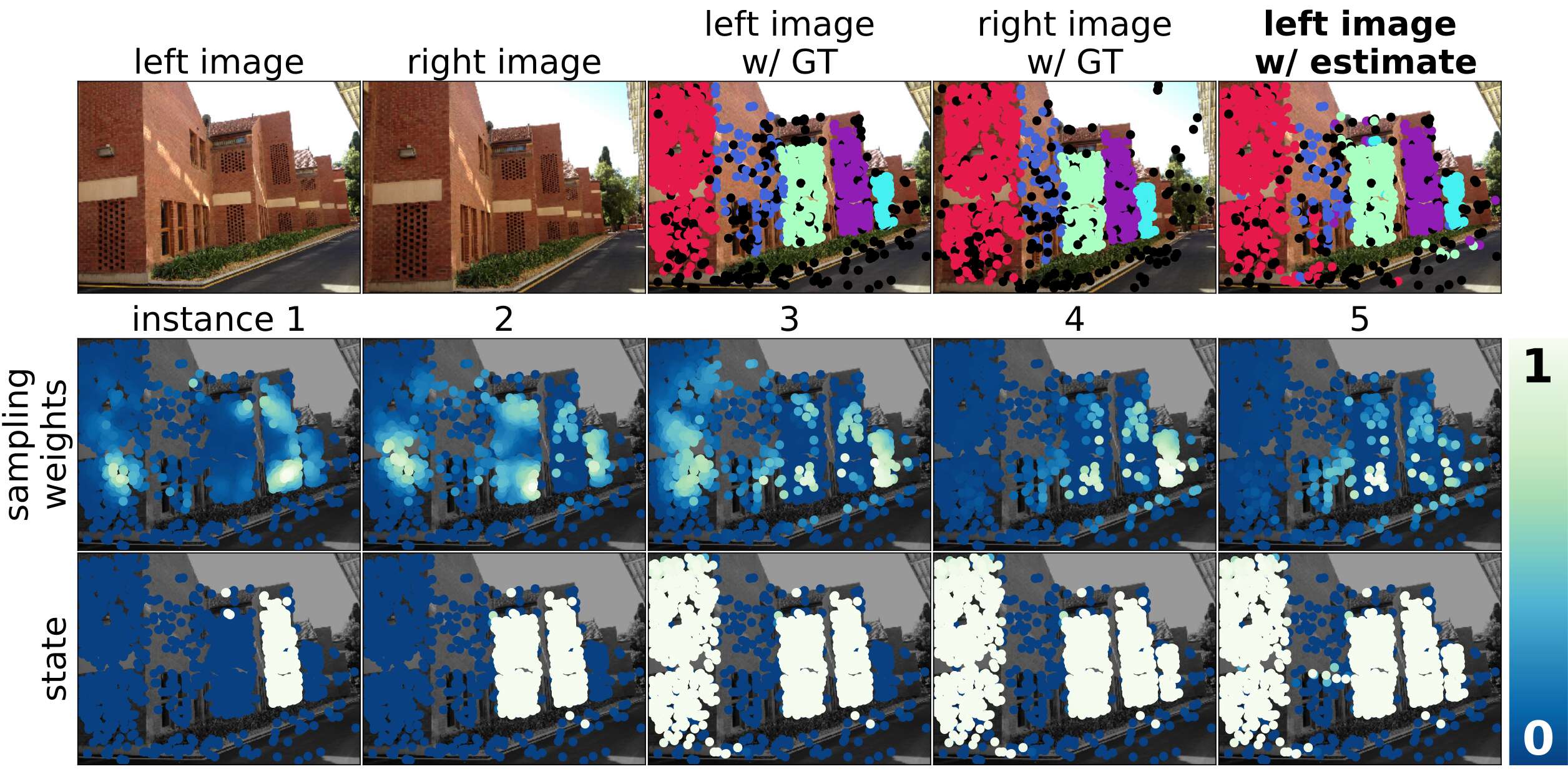}
\caption{\textbf{Homography fitting} result for the AdelaideRMF \emph{unihouse} scene. \textbf{Top:} Left and right image, feature points with ground truth labels, and feature points with labels predicted by CONSAC-S (ME: $8.4\%$). \textbf{Middle:} Sampling weights of feature points at each instance step. \textbf{Bottom:}  State $\vec{s}$ generated from the selected model instances. 
}
\label{fig:adelaide_result_example}
\vspace{-.3em}
\end{figure}

As the authors of \cite{magri2019fitting} used a different evaluation protocol, we recomputed results for MCT using the code provided by the authors. 
For Sequential RANSAC, we used our own implementation. Other results were carried over from~\cite{barath2019progressive} and are shown in Tab.~\ref{tab:hom_results}. 
CONSAC-S outperforms state-of-the-art Progressive-X, yielding a significantly lower average ME with a marginally higher standard deviation. Notably, Sequential RANSAC performs favourably on this task as well. Fig.~\ref{fig:adelaide_result_example} shows a qualitative result for CONSAC-S.

\begin{table}
	\begin{center}
	\begin{tabular}{|l|cc|}
	\cline{2-3}
	\multicolumn{1}{c|}{} & \multicolumn{2}{c|}{AdelaideRMF-H~\cite{wong2011dynamic}}  \\
	\cline{2-3}
	\multicolumn{1}{c|}{} &   \small{avg.} & \small{std.} \\
	\hline
	\textbf{CONSAC-S}                                   & $\mathbf{5.21}$  & $6.46$ \\
	Progressive-X~\cite{barath2019progressive}\tssstar  & $6.86$  & $\mathbf{5.91}$  \\
	Multi-X~\cite{barath2018multi}\tssstar              & $8.71$  & $8.13$ \\
	Sequential RANSAC                                   & $11.14$ & $10.54$ \\
	PEARL~\cite{isack2012energy}\tssstar                & $15.14$ & $6.75$ \\
	MCT~\cite{magri2019fitting}\tssdagger               & $16.21$ & $10.76$ \\
	RPA~\cite{magri2015robust}\tssstar                  & $23.54$ & $13.42$ \\
	T-Linkage~\cite{magri2014t}\tssstar                 & $54.79$ & $22.17$ \\
	RansaCov~\cite{magri2016multiple}\tssstar           & $66.88$ & $18.44$ \\
	\hline
	\end{tabular}
    \end{center}
    \vspace{-4mm}
	\caption{\textbf{Homography estimation:} Average misclassification errors (avg., in \%, lower is better) and their standard deviations (std.) over five runs for homography fitting on the AdelaideRMF~\cite{wong2011dynamic} dataset. *~Results taken from~\cite{barath2019progressive}. \textdagger~Results computed using code provided by the authors. 
	}
	\label{tab:hom_results}
\vspace{-.2mm}
\end{table}

\subsection{Ablation Study}
\label{subsec:ablation}
We perform ablation experiments in order to highlight the effectiveness of several methodological choices. 
As Tab.~\ref{tab:ablation_results} shows, CONSAC with EM refinement consistently performs best on both vanishing point and homography estimation. 
If we disable EM refinement, accuracy drops measurably, yet remains on par with state-of-the-art (cf. Tab.~\ref{tab:vp_results} and Tab.~\ref{tab:hom_results}). 
On NYU-VP we can observe that the self-supervised trained CONSAC-S achieves state-of-the-art performance, but is still surpassed by CONSAC trained in a supervised fashion. 
Training CONSAC-S without inlier masking regularisation (IMR, cf. Sec.~\ref{subsubsec:self_supervised}) reduces accuracy measurably, while training only with IMR and disabling the self-supervised loss produces poor results. 
Switching to unconditional sampling for CONSAC (NYU-VP) or CONSAC-S (AdelaideRMF) comes with a significant drop in performance, and is akin to incorporating vanilla NG-RANSAC~\cite{brachmann2019neural} into Sequential RANSAC. 

\begin{table}
	\begin{center}
	\begin{tabular}{|l|cc|cc|}
	\cline{2-5}
	\multicolumn{1}{c|}{} &   \multicolumn{2}{c|}{NYU-VP} & \multicolumn{2}{c|}{Adelaide} \\
	\cline{2-5}
	\multicolumn{1}{c|}{} &   \small{avg.} & \small{std.} & \small{avg.} & \small{std.} \\
	\hline
	\multicolumn{5}{|c|}{\small{with EM refinement}} \\
	{CONSAC}          & $65.01$ & $0.46$ & --- & ---\\
	{CONSAC-S}        & $63.44$ & $0.40$ & $5.21$ & $6.46$\\
	\hline
	\multicolumn{5}{|c|}{\small{without EM refinement}} \\
	{CONSAC }               & $62.90$ & $0.52$ & --- & ---\\
	{CONSAC-S }             & $61.83$ & $0.58$ & $6.17$ & $7.79$\\
	{CONSAC-S w/o IMR}     & $59.94$ & $0.47$ & $8.14$ & $11.79$\\
	{CONSAC-S only IMR}    & $29.31$ & $0.37$ & $21.12$ & $13.45$\\
	{CONSAC(-S) uncond.}   & $48.36$ & $0.29$ & $9.17$ & $11.50$\\
	\hline
	\end{tabular}
    \end{center}
\vspace{-3mm}
\caption{\textbf{Ablation study:} We compute mean AUC (NYU-VP), mean ME (AdelaideRMF~\cite{wong2011dynamic}) and standard deviations for variations of CONSAC. See Sec.~\ref{subsec:ablation} for details.
	}
	\label{tab:ablation_results}
\vspace{-2mm}
\end{table}

\section{Conclusion}
We have presented CONSAC, the first learning-based robust estimator for detecting multiple parametric models in the presence of noise and outliers. 
A neural network learns to guide model hypothesis selection to different subsets of the data, finding model instances sequentially.
We have applied CONSAC to vanishing point estimation, and multi-homography estimation, achieving state-of-the-art accuracy for both tasks.
We contribute a new dataset for vanishing point estimation which facilitates supervised learning of multi-model estimators, other than CONSAC, in the future. 
\paragraph{Acknowledgements}
This work was supported by the DFG grant \emph{COVMAP} (RO 4804/2-1 and RO 2497/12-2) and has received funding from the European Research Council (ERC)
under the European Union Horizon 2020 programme (grant No. 647769).
\renewcommand*\appendixpagename{\Large Appendix}

\begin{appendices}

This appendix contains additional implementation details (Sec.~\ref{sec:implementation}) which may be helpful for reproducing our results. 
Sec.~\ref{sec:dataset} provides additional details about the datasets presented and used in our paper.
In Sec.~\ref{sec:experiments}, we show additional details complementing our experiments shown in the paper.

\section{Implementation Details}
\label{sec:implementation}
In Alg.~\ref{algo:consac}, we present the CONSAC algorithm in another form, in addition to the description in Sec. 3 of the main paper, for ease of understanding. A list of all user definable parameters and the settings we used in our experiments is given in Tab.~\ref{tab:parameters}.

\begin{algorithm}
\caption{CONSAC}
\label{algo:consac}

\KwIn{$\set{Y}$ -- set of observations, $\vec{w}$ -- network parameters}
\KwOut{$\set{\hat{M}}$ -- multi-hypothesis}

$\set{P} \leftarrow \varnothing$ \;

\For{$i\leftarrow 1$ \KwTo $P$}{
    $\set{M} \leftarrow \varnothing$ \;
    
    $\vec{s} \leftarrow \vec{0}$ \;
    
    \For{$m\leftarrow 1$ \KwTo $M$}{
        $\set{H} \leftarrow \varnothing$ \;
        
        \For{$s\leftarrow 1$ \KwTo $S$}{
            Sample a minimal set of observations $\{\vec{y}_1, \dots, \vec{y}_C\}$ with $\vec{y} \sim p(\vec{y} | \vec{s}; \vec{w})$. \;
            
            $\vec{h} \leftarrow f_{\mathsf{S}}(\{\vec{y}_1, \dots, \vec{y}_C\})$ \;
            
            $\set{H} \leftarrow \set{H} \cup \{\vec{h}\} $ \;
        }
        
        $\vec{\hat{h}} \leftarrow \argmax_{\vec{h} \in \set{H}} g_{\mathsf{s}}(\vec{h}, \set{Y}, \set{M})$ \;
        
        $\set{M} \leftarrow \set{M} \cup \{\vec{\hat{h}}\} $ \;
        
            $\vec{s} \leftarrow \max_{\hat{\vec{h}} \in \set{M}} g_{\mathsf{y}}(\set{Y}, \hat{\vec{h}})$ \;

    }
    
     $\set{P} \leftarrow \set{P} \cup \{\set{M}\} $ \;
    
}

$\hat{\set{M}} \leftarrow \argmax_{\set{M} \in \set{P}} g_{\mathsf{m}}(\set{M}, \set{Y})$ \;
% \vspace{1em}
\end{algorithm}

\begin{table}[t]
\setlength\tabcolsep{.30em}
    \centering
    \begin{tabular}{c|lc|c|c|}
        \multicolumn{3}{c|}{}                           & VP         & homography \\
        \multicolumn{3}{c|}{}                           & estimation & estimation \\
        \hline
        \parbox[t]{2.5mm}{\multirow{11}{*}{\rotatebox[origin=c]{90}{training}}}
        &  learning rate & & $10^{-4}$ & $2\cdot10^{-6}$ \\
        &  batch size & $B$ & $16$ & $1$ \\
        &  \multicolumn{2}{l|}{batch normalisation} & yes & no \\
        &  epochs & & $400$ & $100$ \\
        &  inlier threshold & $\tau$  & $10^{-3}$ & $10^{-4}$  \\
        &  IMR weight & $\kappa$  & $10^{-2}$ & $10^{-2}$  \\
        &  observations per scene & $|\set{Y}|$ & $256$ & $256$ \\
        &  number of instances & $M$  & $3$ & $6$  \\
        &  single-instance samples & $S$  & $2$ & $2$  \\
        &  multi-instance samples & $P$  & $2$ & $2$  \\
        &  sample count & $K$  & $4$ & $8$  \\
        \hline
        \parbox[t]{2.5mm}{\multirow{9}{*}{\rotatebox[origin=c]{90}{test}}}
        &  inlier threshold & $\tau$  & $10^{-3}$ & $10^{-4}$  \\
        &  inlier thresh. (selection) & $\theta$  & --- & $3\cdot10^{-3}$  \\
        &  inlier cutoff (selection) & $\Theta$  & --- & $6$  \\
        &  observations per scene & $|\set{Y}|$ & \multicolumn{2}{c|}{variable}\\
        &  number of instances & $M$  & $6$ & $6$  \\
        &  single-instance samples & $S$  & $32$ & $100$  \\
        &  multi-instance samples & $P$  & $32$ & $100$  \\
        &  EM iterations &   & $10$ & $10$  \\
        &  EM standard deviation & $\sigma$  & $10^{-8}$ & $10^{-9}$  \\
        
    \end{tabular}
    \caption{\textbf{User definable parameters} of CONSAC and the values we chose for our experiments on vanishing point estimation and homography estimation. We distinguish between values used during training and at test time. Mathematical symbols refer to the notation used either in the main paper or in this supplementary document.  }
    \label{tab:parameters}
    % \vspace{-1em}
\end{table}

\subsection{Neural Network}
\begin{figure*}
\centering
\includegraphics[width=0.99\linewidth]{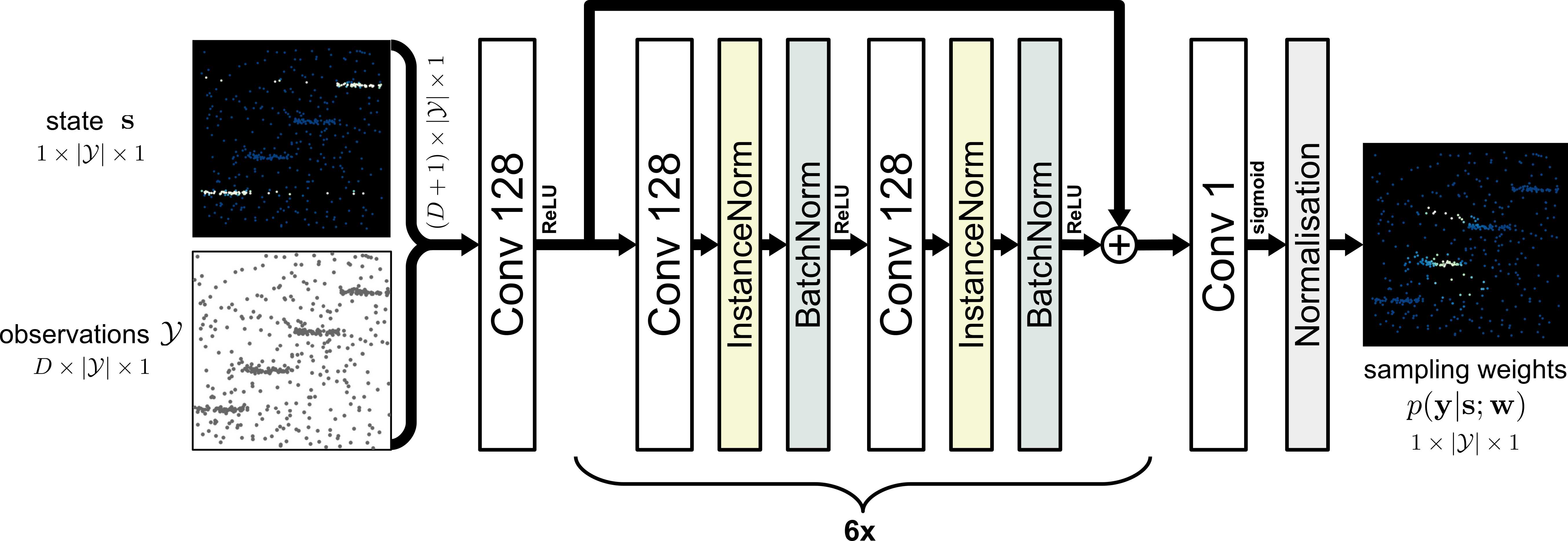}
\caption{\textbf{CONSAC neural network architecture} used for all experiments. 
We stack observations $\set{Y}$, \eg line segments or point correspondences (\emph{not} an image), and state $\vec{s}$ into a tensor of size $(D+1) \times |\set{Y}| \times 1$, and feed it into the network. The network is composed of linear $1\times 1$ convolutional layers interleaved with instance normalisation \cite{ulyanov2016instance}, batch normalisation \cite{ioffe2015batch} and ReLU \cite{he2015delving} layers which are arranged as residual blocks~\cite{he2016deep}. Only using $1 \times 1$ convolutions, the network is order invariant \wrt observations $\set{Y}$. The architecture is based on \cite{brachmann2019neural,goodcorr18}. }
\label{fig:cnn_arch}
\end{figure*}   
We use a neural network similar to PointNet~\cite{qi2017pointnet} and based on~\cite{brachmann2019neural, goodcorr18} for prediction of conditional sampling weights in CONSAC. 
Fig.~\ref{fig:cnn_arch} gives an overview of the architecture. 
Observations $\vec{y} \in \set{Y}$, \eg line segments or feature point correspondences, are stacked into a tensor of size $D \times |\set{Y}| \times 1$. 
Note that the size of the tensor depends on the number of observations per scene. 
The dimensionality $D$ of each observation $\vec{y}$ is application specific. 
The current state $\vec{s}$ contains a scalar value for each observation and is hence a tensor of size  $1 \times |\set{Y}| \times 1$. 
The input of the network is a concatenation of observations $\set{Y}$ and state $\vec{s}$, \ie a tensor of size $(D+1) \times |\set{Y}| \times 1$.  
After a single convolutional layer ($1\times1$, $128$ channels) with ReLU~\cite{he2015delving} activation function, we apply six residual blocks~\cite{he2016deep}. 
Each residual block is composed of two series of convolutions ($1\times1$, $128$ channels), instance normalisation \cite{ulyanov2016instance}, batch normalisation \cite{ioffe2015batch} (optional) and ReLU activation.
After another convolutional layer ($1\times1$, $1$ channel) with sigmoid activation, we normalise the outputs so that the sum of sampling weights equals one. 
Only using $1 \times 1$ convolutions, this network architecture is order invariant \wrt observations $\set{Y}$. 
We implement the architecture using PyTorch~\cite{paszke2017automatic} version 1.2.0.

\subsubsection{Training Procedure}
We train the neural network using the Adam~\cite{kingma2014adam} optimiser and utilise a cosine annealing learning rate schedule~\cite{loshchilov2016sgdr}. 
We clamp losses to a maximum absolute value of $0.3$ in order to avoid divergence caused by large gradients resulting from large losses induced by poor hypothesis samples.

\paragraph{Number of Observations} 
In order to keep the number of observations $|\set{Y}|$ constant throughout a batch, we sample a fixed number of observations from all observations of a scene during training. 
At test time, all observations are used.

\paragraph{Pseudo Batches}
During training, we sample $P$ multi-hypotheses $\set{M}$, from which we select the best multi-hypothesis $\hat{\set{M}}$ for each set of input observations $\set{Y}$ within a batch of size $B$. 
To approximate the expectation of our training loss (see Sec.~3.2 of the main paper), we repeat this process $K$ times, to generate $K$ samples of \emph{selected} multi-hypotheses $\hat{\set{M}}$ for each $\set{Y}$. 
We generate each multi-hypothesis $\set{M}$ by \emph{sequentially} sampling $S$ single-instance hypotheses $\vec{h}$ and selecting the best one, conditioned on a state $\vec{s}$.
The state $\vec{s}$ varies between these innermost sampling loops, since we compute $\vec{s}$ based on all previously selected single instance hypotheses $\hat{\vec{h}}$ of a multi-hypothesis $\set{M}$.
Because $\vec{s}$ is always fed into the network alongside observations $\set{Y}$, we have to run $P\cdot K$ forward passes for each batch. 
We can, however, parallelise these passes by collating observations and states into a tensor of size $P \times K \times B \times (D+1) \times |\set{Y}|$. 
We reshape this tensor so that it has size  $B^* \times (D+1) \times |\set{Y}|$ with an effective pseudo batch size $B^* = P \cdot K \cdot B$, in order to process all samples in parallel while using the same neural network weights for each pass within $B^*$. 
This means that sample sizes $P$ and $K$ are subject to both time and hardware memory constraints.
We observe, however, that small sample sizes during training are sufficient in order to achieve good results using higher sample sizes at test time. 

\paragraph{Inlier Masking Regularisation} For self-supervised training, we multiply the inlier masking regularisation (IMR) term $\ell_{\mathsf{im}}$ (cf. Sec. 3.2.2 in the main paper) with a factor $\kappa$ in order to regulate its influence compared to the regular self-supervision loss $\ell_{\mathsf{self}}$, \ie:
\begin{equation}
    \ell = \ell_{\mathsf{self}} + \kappa \cdot \ell_{\mathsf{im}}
\end{equation}

\subsection{Scoring Functions}

In order to gauge whether an observation $\vec{y}$ is an inlier of model instance $\vec{h}$, we utilise a soft inlier function adapted from~\cite{brachmann2018lessmore}:
\begin{equation}
    g_{\mathsf{i}}(\vec{y}, \vec{h}) = 1-\sigma(\beta r(\vec{y}, \vec{h}) - \beta \tau) \, ,
\end{equation}
with inlier threshold $\tau$, softness parameter $\beta = 5\tau^{-1}$, a task-specific residual function $r(\vec{y}, \vec{h})$ (see Sec.~\ref{subsec:residual_fun} for details), and using the sigmoid function:
\begin{equation}
    \sigma(x) = \frac{1}{1+e^{-x}} \, .
\end{equation}

The multi-instance scoring function $g_{\mathsf{m}}$, which we use to select the best muti-hypothesis, \ie hypothesis of multiple model instances $\set{\hat{M}} = \{\vec{\hat{h}}_1, \dots, \vec{\hat{h}}_M\}$, from a pool of multi-instance hypotheses $\set{P} = \{ \set{M}_1, \dots, \set{M}_P \}$, counts the joint inliers of all models in a multi-instance:
\begin{equation}
g_{\mathsf{m}}(\set{M}, \set{Y}) = \sum_{\vec{y} \in \set{Y}} \max_{{\vec{h}} \in \set{M}} g_{\mathsf{i}}(\vec{y}, \vec{{h}})  \, .
\end{equation}

The single instance scoring function $g_{\mathsf{s}}$, which we use for selection of single model instances $\vec{h}$ given the set of previously selected model instances $\set{M}$, is a special case of the multi-instance scoring function $g_{\mathsf{m}}$:
\begin{equation}
g_{\mathsf{s}}(\vec{h}, \set{Y}, \set{M}) = g_{\mathsf{m}}(\set{M} \cup \{\vec{h}\}, \set{Y})  \, .
\end{equation}

\subsection{Residual Functions}
\label{subsec:residual_fun}
\paragraph{Line Fitting}
For the line fitting problem, each observation is a 2D point in homogeneous coordinates $\vec{y} = \left(x\, y\, 1\right)\tran$, and each model is a line in homogeneous coordinates $\vec{h} = \frac{1}{\|(n_1\,n_2)\|}\left(n_1\, n_2\, d\right)\tran$. 
We use the absolute point-to-line distance as the residual:
\begin{equation}
    r(\vec{y}, \vec{h}) = |\vec{y}\tran\vec{h}| \, .
\end{equation}

\paragraph{Vanishing Point Estimation}

\begin{figure}	
\centering
\includegraphics[width=0.95\linewidth]{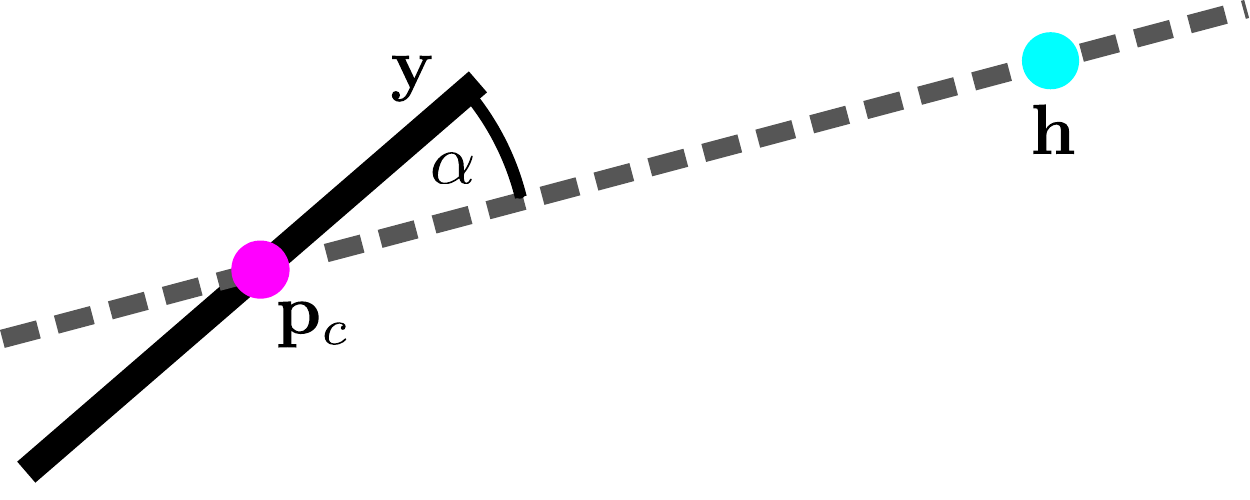}
\caption{Visualisation of the angle $\alpha$ used for the vanishing point estimation residual function $r(\vec{y}, \vec{h})$.}
\label{fig:consistency_angle}
\end{figure}

Observations $\vec{y}$ are given by line segments with start point $\vec{p}_1 = \left(x_1 \, y_1 \, 1\right)\tran$ and end point $\vec{p}_2 = \left(x_2 \, y_2 \, 1\right)\tran$, and models are vanishing points $\vec{h} = \left(x\, y\, 1\right)\tran$. 
For each line segment $\vec{y}$, we compute the corresponding line $\vec{l}_{\vec{y}} = \vec{p}_1 \times \vec{p}_2$ and the centre point $\vec{p}_c = \frac{1}{2} (\vec{p}_1 + \vec{p}_2)$. 
As visualised by Fig.~\ref{fig:consistency_angle}, we define the residual via the cosine of the angle $\alpha$ between $\vec{l}_{\vec{y}}$ and the constrained line $\vec{l}_c = \vec{h} \times \vec{p}_c$, \ie the line connecting the vanishing point with the centre of the line segment:
\begin{equation}
    r(\vec{y}, \vec{h}) = 1-\cos \alpha =  1- \frac{|\vec{l}_{\vec{y},1:2}\tran \vec{l}_{c,1:2} |}{\|\vec{l}_{\vec{y},1:2}\| \|\vec{l}_{c,1:2}\|}    \, .
\end{equation}

\paragraph{Homography Estimation}
Observations $\vec{y}$ are given by point correspondences $\vec{p}_1 = \left(x_1 \, y_1 \, 1\right)\tran$ and $\vec{p}_2 = \left(x_2 \, y_2 \, 1\right)\tran$, and models are plane homographies $\vec{h} = \vec{H}^{3\times3}$ which shall map $\vec{p}_1$ to $\vec{p}_2$. 
We compute the symmetric squared transfer error:
\begin{equation}
    r(\vec{y}, \vec{h}) =  \| \vec{p}_1 - \vec{p}_1' \|^2 +  \| \vec{p}_2 - \vec{p}_2' \|^2  \, ,
\end{equation}
with $\vec{p}_2' \propto \vec{H}\vec{p}_1$ and $\vec{p}_1' \propto \vec{H}\inverse\vec{p}_2$.

\section{Dataset Details and Analyses}
\label{sec:dataset}

\subsection{Line Fitting}
\begin{figure}
\centering
\includegraphics[width=0.99\linewidth]{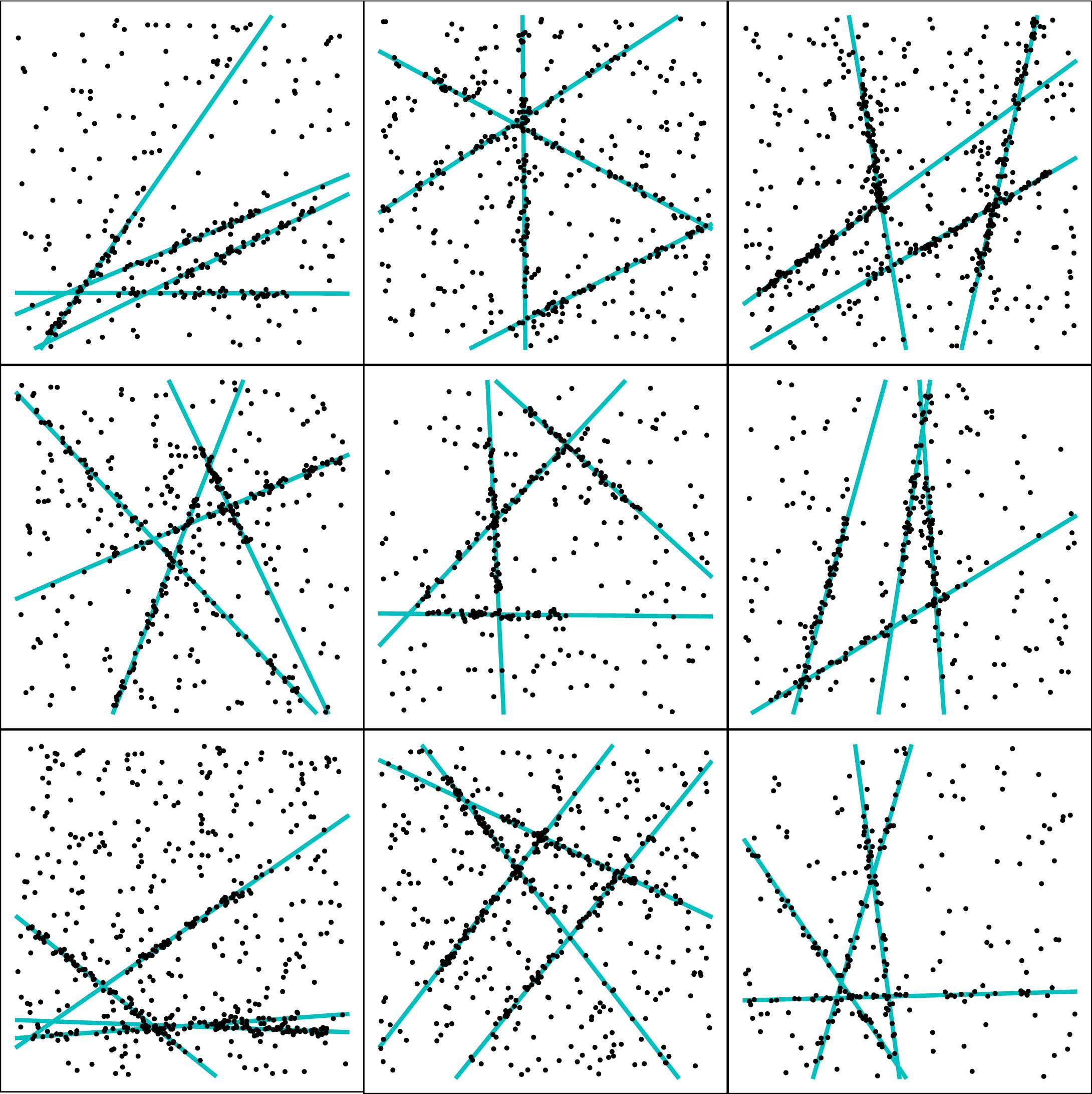}
\caption{\textbf{Line fitting:} we show examples from the synthetic dataset we used to train CONSAC on the line fitting problem. Each scene consists of four lines placed at random, with points sampled along them, perturbed by Gaussian noise and outliers. \textcolor{cyan}{Cyan} = ground truth lines.}
\label{fig:synth_line_dataset_examples}
\end{figure}  

For training CONSAC on the line fitting problem, we generated a synthetic dataset of $10000$ scenes. Each scene consists of four lines placed at random within a $\{0,1\} \times \{0,1\}$ square. For each line, we randomly define a line segment with a length of $30-100\%$ of the maximum length of the line within the square. Then, we randomly sample $40-100$ points along the line segment and perturb them by Gaussian noise $\mathcal{N}\sim(0,\,\sigma^{2})$, with $\sigma \in (0.007, 0.008)$ sampled uniformly. Finally, we add $40-60\%$ outliers via random uniform sampling. Fig.~\ref{fig:synth_line_dataset_examples} shows a few examples from this dataset.

\begin{figure}	
\centering
\includegraphics[width=0.99\linewidth]{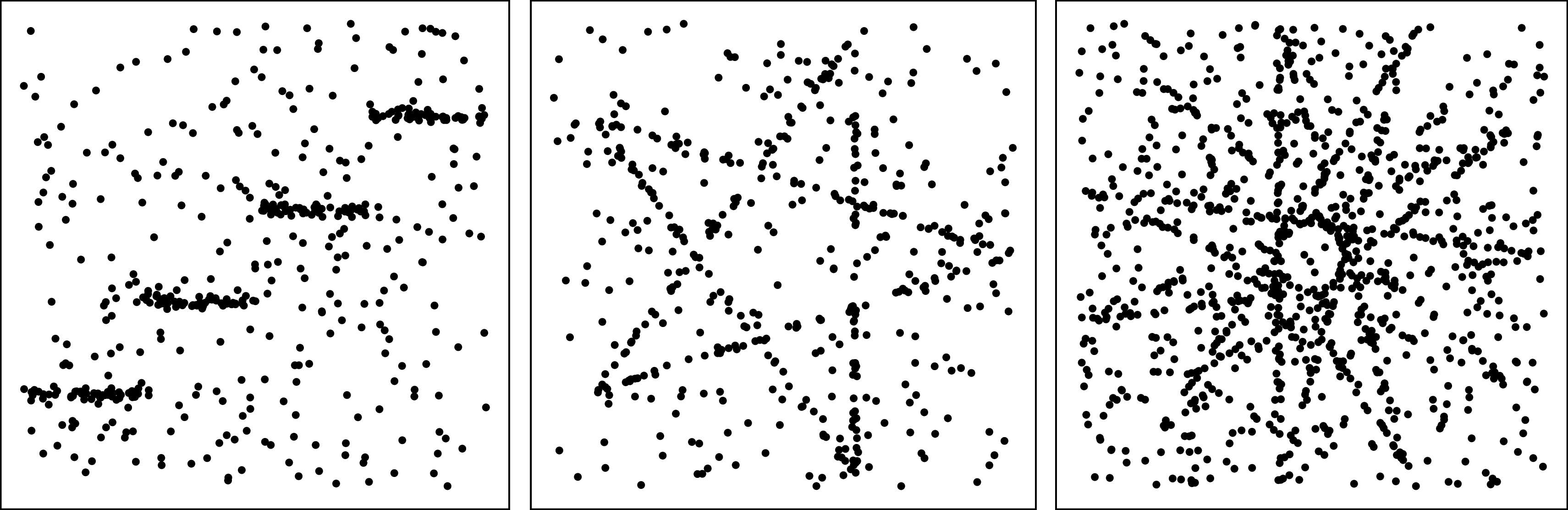}
\caption{\textbf{Line fitting:} we use the synthetic \emph{stair4} (left), \emph{star5} (middle) and \emph{star11} (right) scenes from~\cite{toldo2008robust}, which were also used by~\cite{barath2019progressive}, in our experiments. }
\label{fig:stair_stars}
\end{figure}

For evaluation, we use the synthetic \emph{stair4}, \emph{star5} and \emph{star11} scenes from~\cite{toldo2008robust}, which were also used by~\cite{barath2019progressive}. As Fig.~\ref{fig:stair_stars} shows, each scene consists of 2D points forming four, five or eleven line segments. The points are perturbed by Gaussian noise ($\sigma = 0.0075$) and contain $50-60\%$ outliers.

\subsection{Vanishing Point Estimation}
\begin{figure*}
\centering
\begin{subfigure}[c]{0.99\linewidth}
\includegraphics[width=0.99\linewidth]{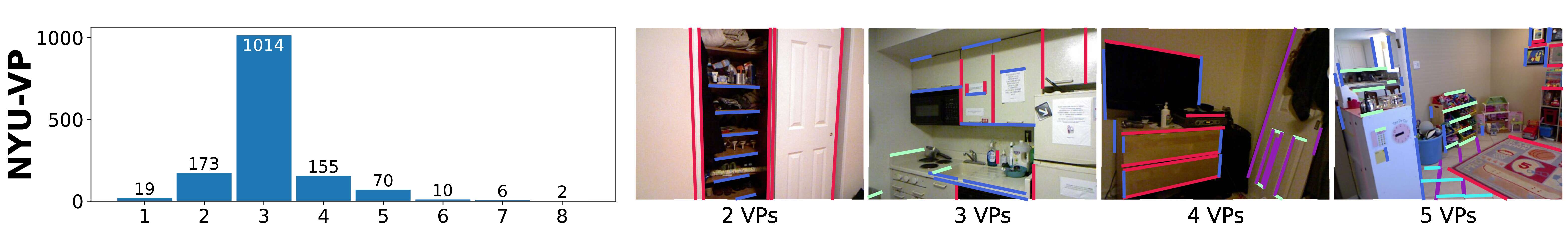}
\label{fig:vp_hist_nyu}
\end{subfigure}
\begin{subfigure}[c]{0.99\linewidth}
\includegraphics[width=0.99\linewidth]{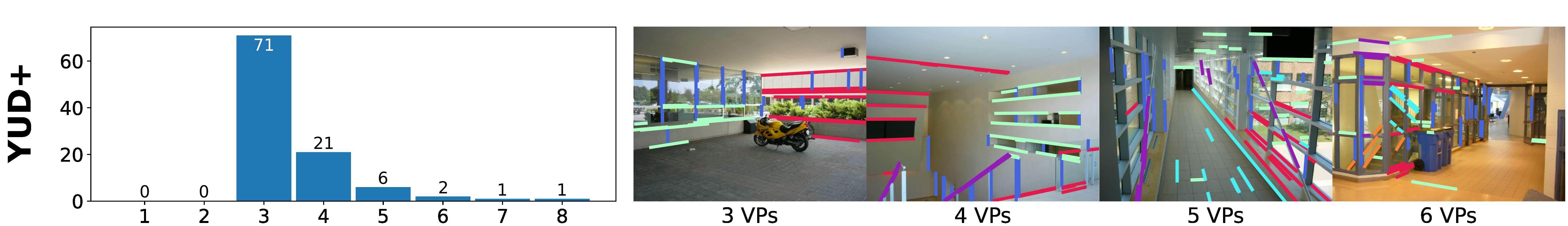}
\label{fig:vp_hist_yudplus}
\end{subfigure}
\caption{\textbf{Vanishing points per scene:} Histograms showing the numbers of vanishing point instances per image for our new NYU-VP dataset (top) and our YUD+ dataset extension (bottom), in addition to a few example images. We illustrate the vanishing points present in each example via colour-coded line segments.}
\label{fig:vp_hists}
\end{figure*}   
% \noindent
\paragraph{NYU-VP} In Fig.~\ref{fig:vp_hists} (top), we show a histogram of the number of vanishing points per image in our new NYU-VP dataset. In addition, we show a few example images for different numbers of vanishing points. NYU-VP solely consists of indoor scenes. 
\paragraph{YUD+} In Fig.~\ref{fig:vp_hists} (bottom), we show a histogram of the number of vanishing points per image in our new YUD+ dataset extension. By comparison, the original YUD~\cite{denis2008efficient} contains exactly three vanishing point labels for each of the $102$ scenes. YUD contains both indoor and outdoor scenes.

\subsection{Homography Estimation}
For self-supervised training for the task of homography estimation, we use SIFT \cite{Lowesift} feature correspondences extracted from the structure-from-motion scenes of~\cite{reconworld15,strecha2008benchmarking,xiao2013sun3d}. Specifically, we used the outdoor scenes \emph{Buckingham}, \emph{Notredame}, \emph{Sacre Coeur}, \emph{St.~Peter's} and \emph{Reichstag} from \cite{reconworld15}, \emph{Fountain} and \emph{Herzjesu} from \cite{strecha2008benchmarking}, and $16$ indoor scenes from SUN3D~\cite{xiao2013sun3d}. We use the SIFT correspondences computed and provided by Brachmann and Rother~\cite{brachmann2019neural}, and discard suspected gross outliers with a matching score ratio greater than $0.9$. As this dataset is imbalanced in the sense that some scenes contain significantly more image pairs than others -- for \emph{St.~Peter's} we have $9999$ image pairs, but for \emph{Reichstag} we only have $56$ --  we apply a rebalancing sampling during training: instead of sampling image pairs uniformly at random, we uniformly sample one of the scenes first, and then we sample an image pair from within this scene. This way, each scene is sampled during training at the same rate. During training, we augment the data by randomly flipping all points horizontally or vertically, and shifting and scaling them along both axes independently by up to $\pm 10\%$ of the image width or height.

\section{Additional Experimental Results}
\label{sec:experiments}

\subsection{Line Fitting}
\begin{figure}	
\centering
\includegraphics[width=0.99\linewidth]{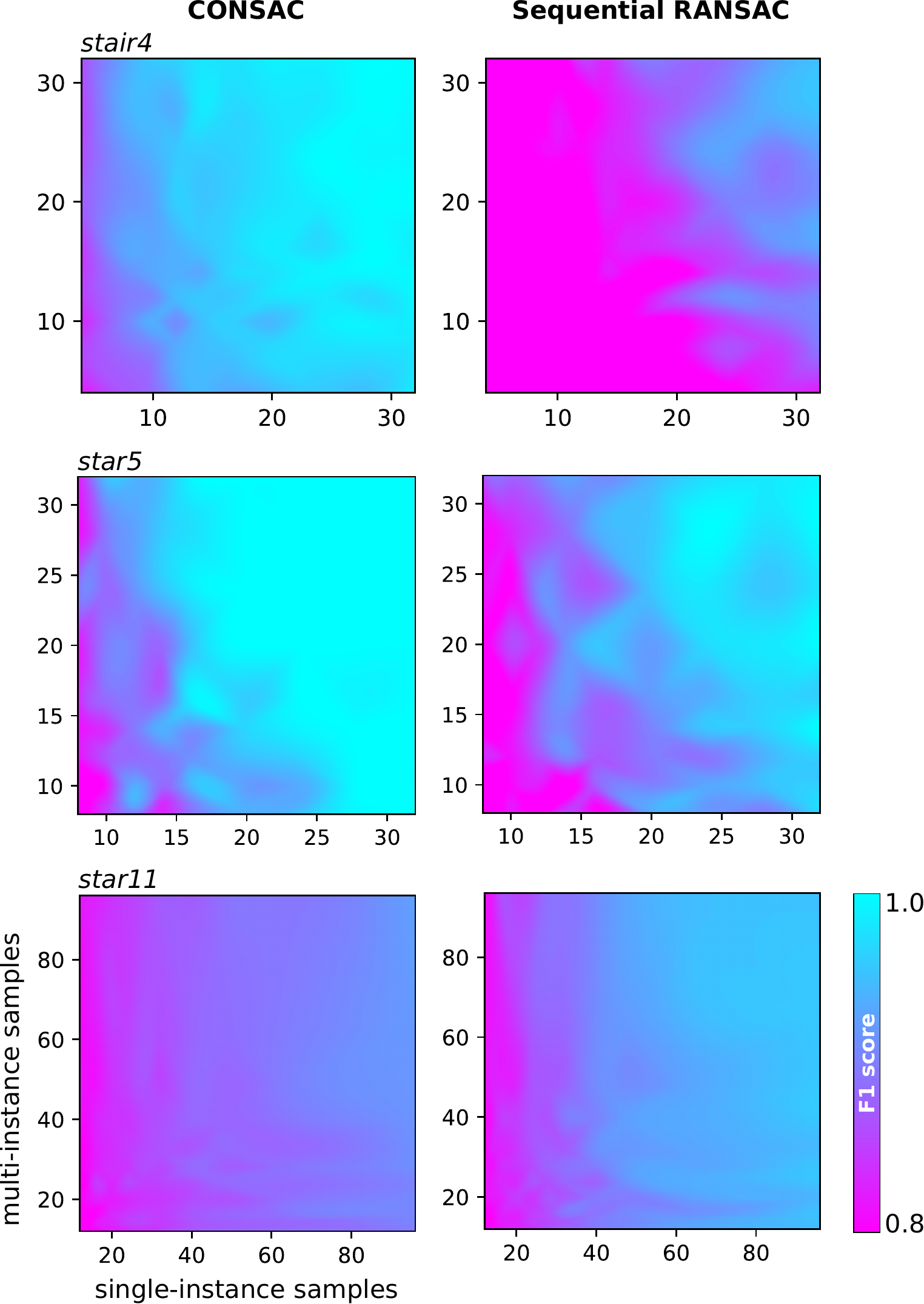}
\caption{\textbf{Line fitting:} Using the \emph{stair4} (top), \emph{star5} (middle) and \emph{star11} (bottom) line fitting scenes from~\cite{toldo2008robust}, we compute the $F1$ scores for various combinations of single-instance samples $S$ (abscissa) and multi-instance samples $P$ (ordinate) and plot them as a heat map. We compare CONSAC (left) with Sequential RANSAC (right). \textcolor{magenta}{Magenta} = low, \textcolor{cyan}{cyan} = high $F1$ score. }
\label{fig:star5_f1_heatmap}
% \vspace{-1em}
\end{figure}

\paragraph{Sampling Efficiency}
In order to analyse the efficiency of the conditional sampling of CONSAC compared to a Sequential RANSAC, we computed the $F1$ score \wrt estimated model instances on the \emph{stair4}, \emph{star5} and \emph{star11} line fitting scenes from~\cite{toldo2008robust} for various combinations of single-instance samples $S$ and multi-instance samples $P$. As Fig.~\ref{fig:star5_f1_heatmap} shows, CONSAC achieves higher $F1$ scores with fewer hypotheses on  \emph{stair4} and  \emph{star5}. As we trained CONSAC on data containing only four line segments, while  \emph{star5} depicts five lines, this demonstrates that CONSAC is able to generalise beyond the number of model instances it has been trained for. On \emph{star11}, which contains eleven lines, it does not perform as well, suggesting that this generalisation may not extend arbitrarily beyond numbers of instances CONSAC has been trained on. In practice, however, our real-world experiments on homography estimation and vanishing point estimation show that it is sufficient to simply train CONSAC on a reasonably large number of instances in order to achieve very good results.

\begin{figure}	
\centering
\includegraphics[width=0.95\linewidth]{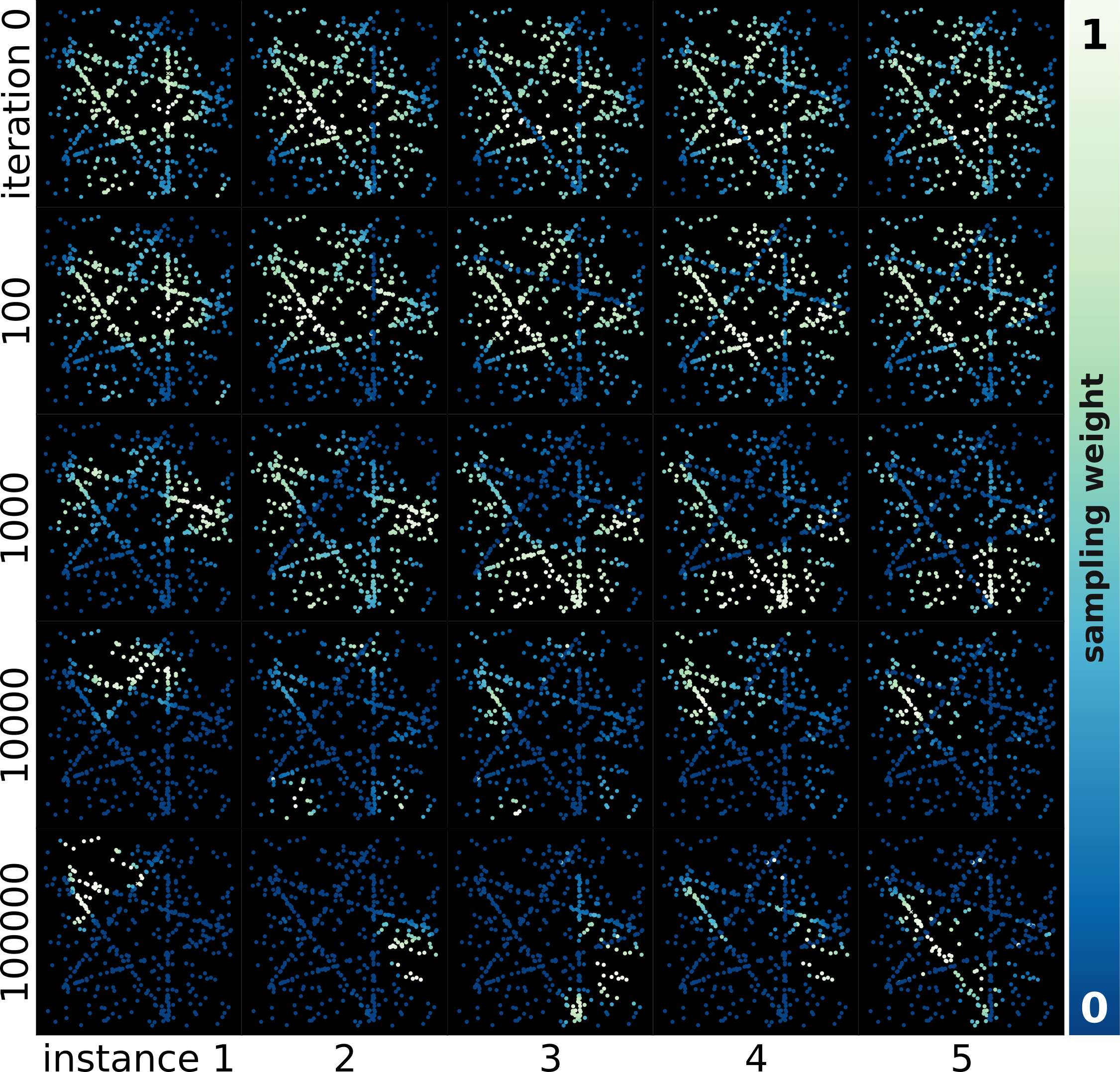}
\caption{\textbf{Line fitting:} We show how the sampling weights at each instance sampling step develop as neural network training progresses, using the \emph{star5} line fitting scene from~\cite{toldo2008robust} as an example. Each row depicts the sampling weights used to sample the eventually selected best multi-hypothesis $\set{\hat{M}}$. \textbf{Top to bottom:} training iterations $0-100000$. \textbf{Left to right:} model instance sampling steps $1-5$. Sampling weights: \textcolor{blue}{Blue} = low, white = high. }
\label{fig:star5_iter}
% \vspace{-1em}
\end{figure}

\noindent
\paragraph{Sampling Weights Throughout Training} 
We looked at the development of sampling weights as neural network training progresses, using \emph{star5} as an example. As Fig.~\ref{fig:star5_iter} shows, sampling weights are randomly -- but not uniformly -- distributed throughout all instance sampling steps before training has begun. At $1000$ iterations, we observe that the neural network starts to focus on different regions of the data throughout the instance sampling steps. From thereon, this focus gets smaller and more accurate as training progresses. After $100000$ iterations, the network has learned to focus on points mostly belonging to just one or two true line segments.

\subsection{Vanishing Point Estimation}
\begin{figure}	
\centering
\begin{subfigure}[c]{\linewidth}
\includegraphics[width=0.99\linewidth]{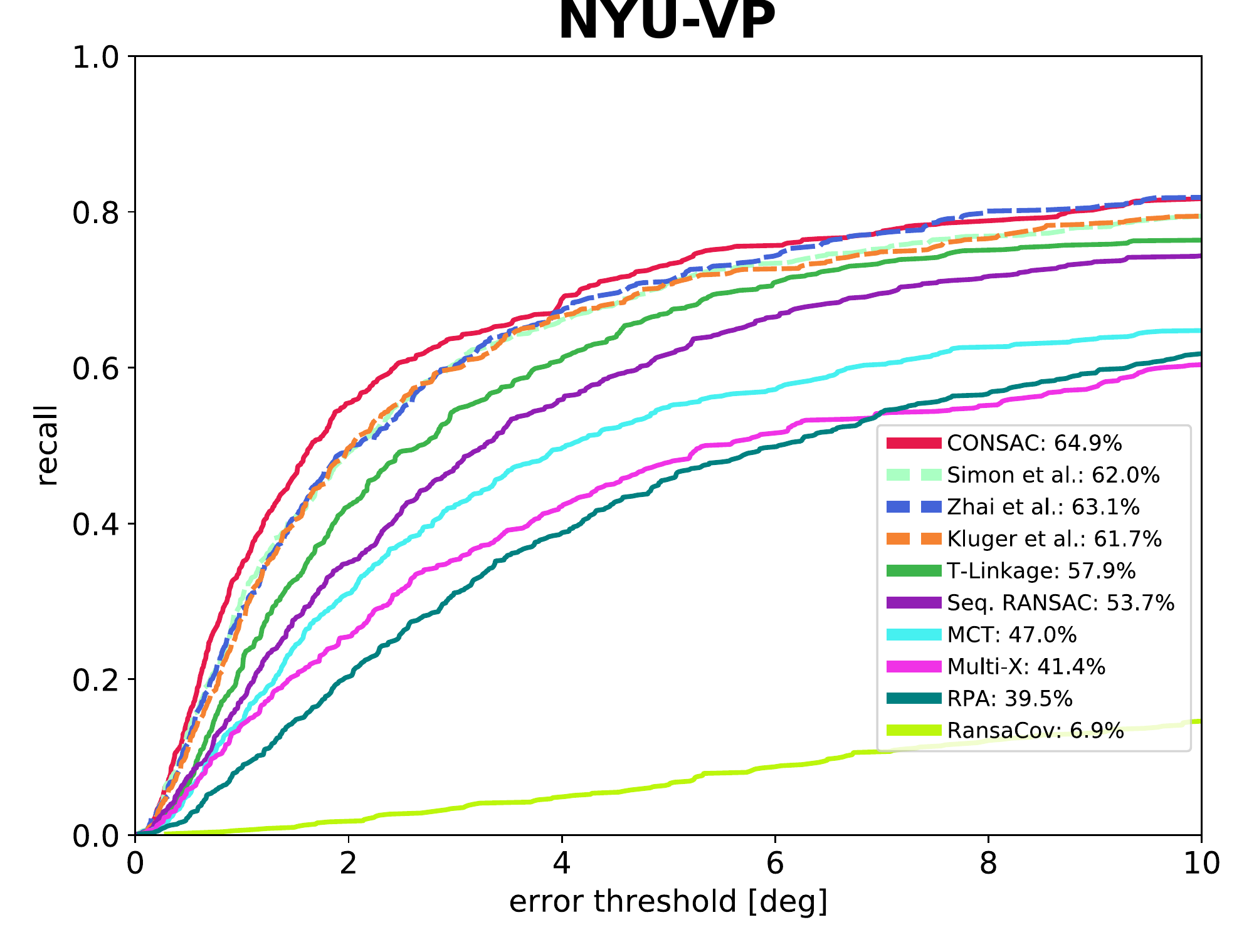}
\vspace{0.3em}
\end{subfigure}
\begin{subfigure}[c]{\linewidth}
\includegraphics[width=0.99\linewidth]{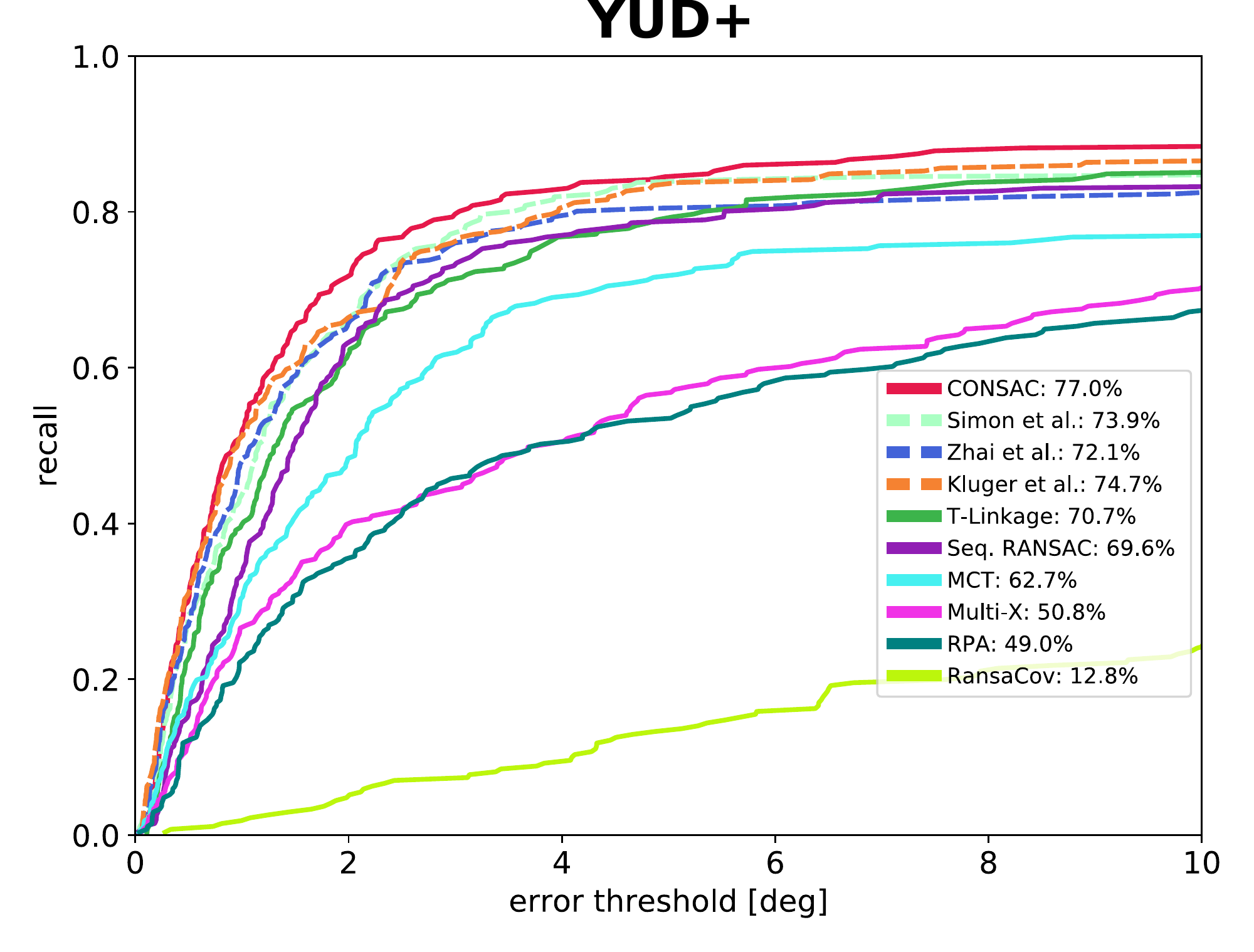}
\vspace{0.3em}
\end{subfigure}
\begin{subfigure}[c]{\linewidth}
\includegraphics[width=0.99\linewidth]{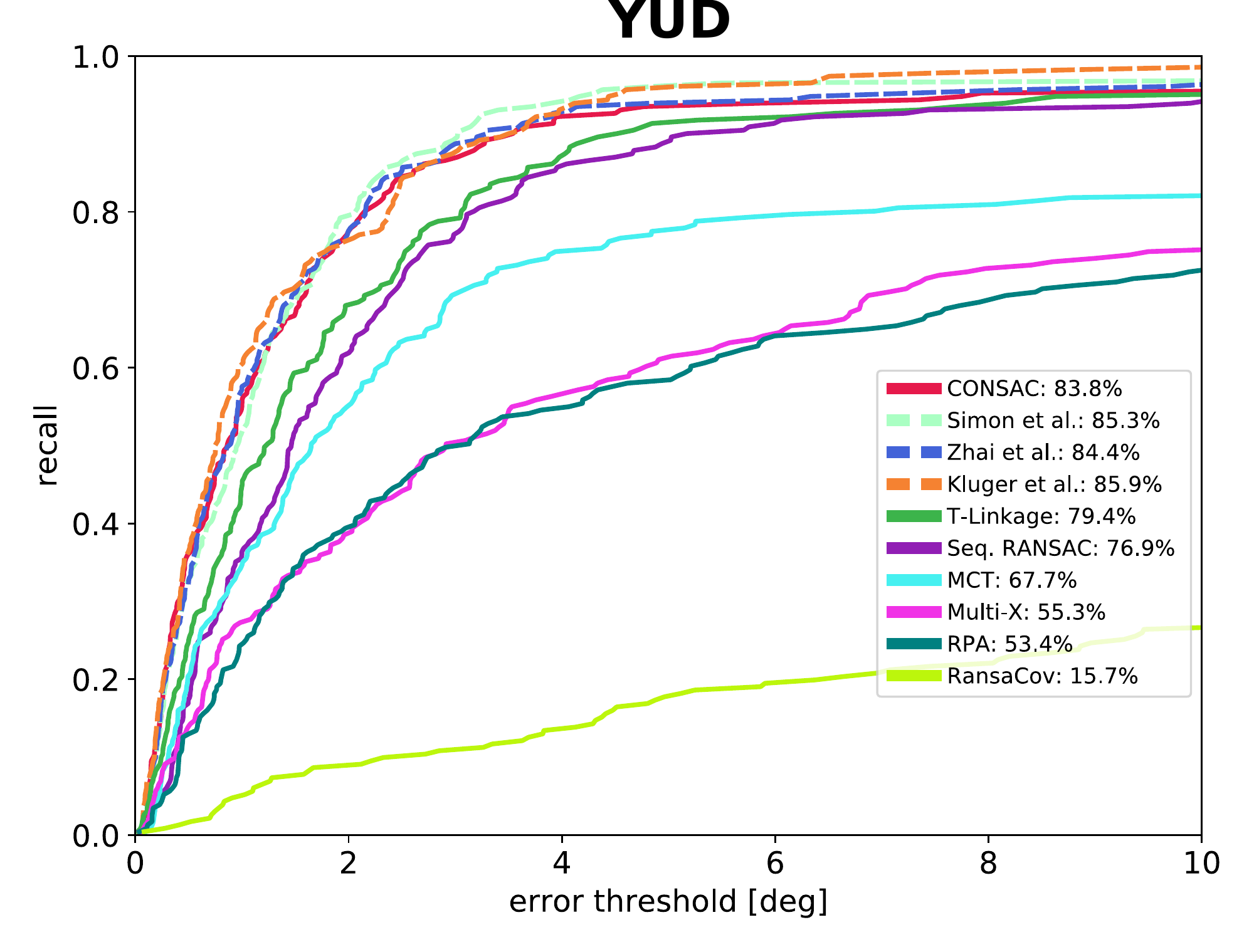}
\end{subfigure}
\caption{\textbf{Vanishing point estimation:} Recall curves for errors up to $10^{\circ}$ for all methods which we considered in our experiments. We selected the result with the \emph{median} AUC out of five runs for each method. Robust estimators are represented with solid lines, task-specific VP estimators with dashed lines. \textbf{Top:} Results on our new NYU-VP dataset. \textbf{Middle:} Results on our new YUD+ dataset extension. \textbf{Bottom:} Results on the original YUD~\cite{denis2008efficient}. }
\label{fig:vp_recall}
\end{figure}
\paragraph{Evaluation Metric}
We denote ground truth VPs of an image by $\set{V} = \{\vec{v}_1, \dots, \vec{v}_M\}$ and estimates by $\set{\hat{V}} = \{\vec{\hat{v}}_1, \dots, \vec{\hat{v}}_N\}$. We compute the error between two particular VP instances via the angle $e(\vec{v}, \vec{\hat{v}})$ between their corresponding directions in 3D using camera intrinsics $\vec{K}$:  

\begin{equation}
    e(\vec{v}, \vec{\hat{v}}) = \arccos{ \frac{ \left| \left(\vec{K}\inverse \vec{v}\right)\tran \vec{K}\inverse \vec{\hat{v}} \right| }{ \left|| \vec{K}\inverse \vec{v} \right|| \cdot \left|| \vec{K}\inverse \vec{\hat{v}} \right|| } } \, .
\end{equation}
We use this error to define the cost matrix $\vec{C}$: $C_{ij} = e(\vec{v}_i, \vec{\hat{v}}_j)$ in Sec.~5.2.1 of the main paper.

\begin{figure*}	
\centering
\vspace{-0.4em}
\begin{subfigure}[c]{\linewidth}
\centering
\includegraphics[width=0.85\linewidth]{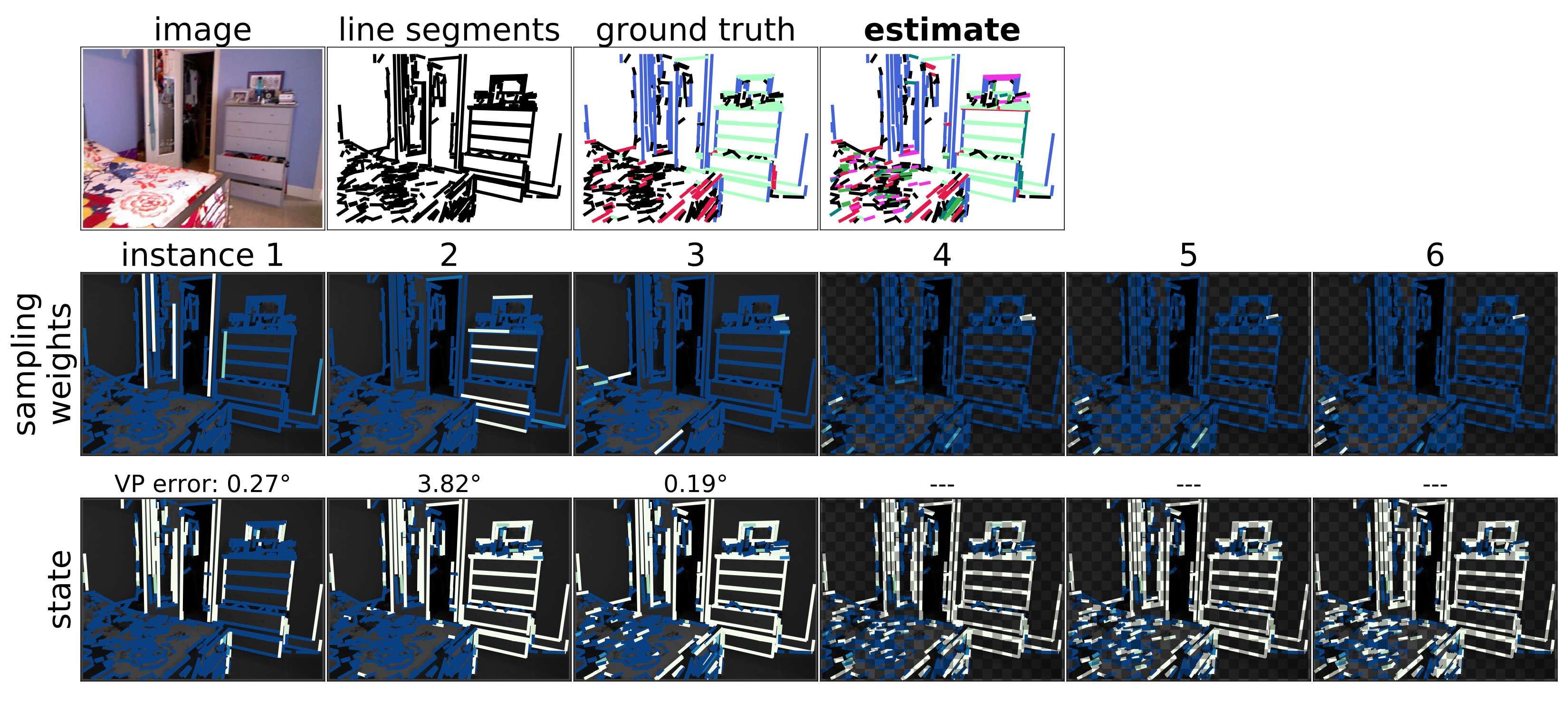}
\vspace{0.2em}
\end{subfigure}
\begin{subfigure}[c]{\linewidth}
\centering
\includegraphics[width=0.85\linewidth]{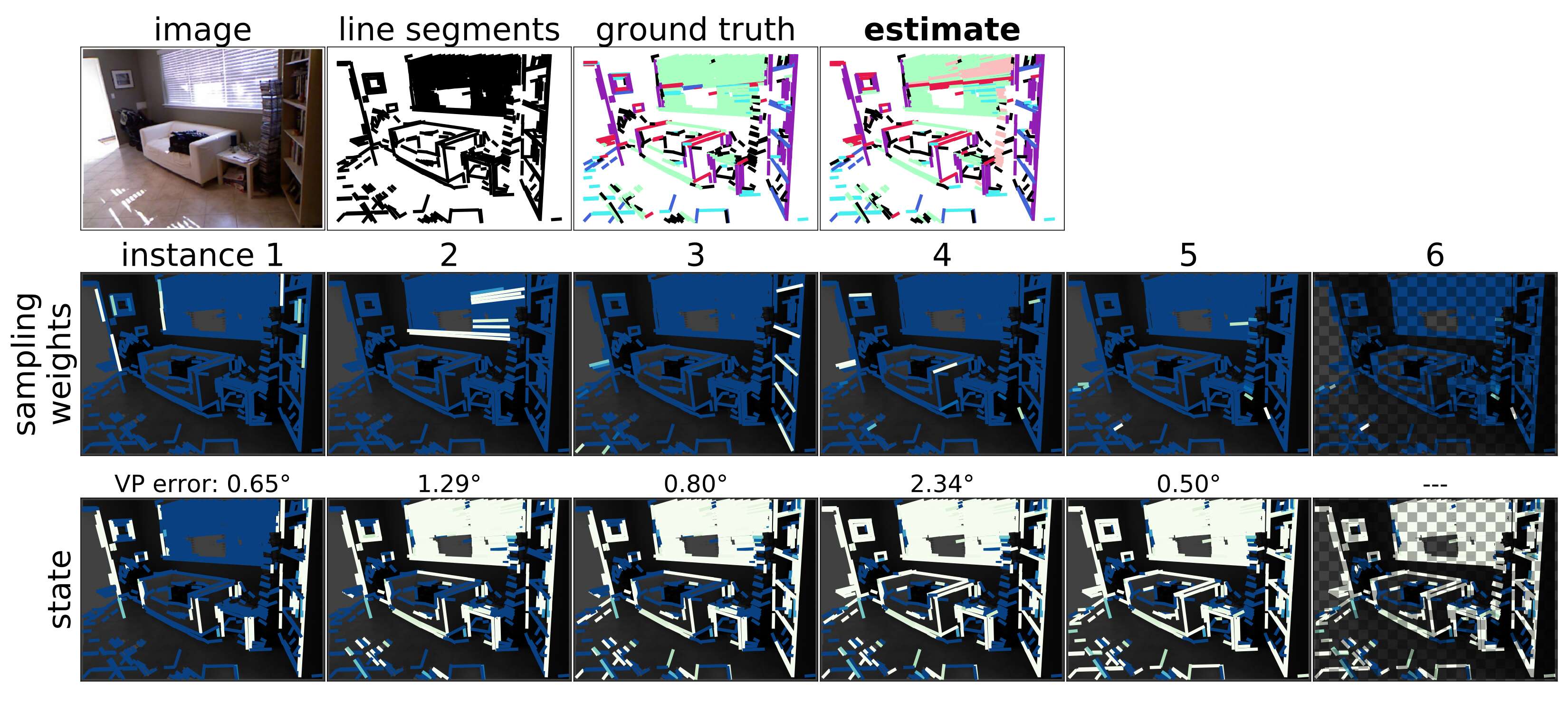}
\vspace{0.2em}
\end{subfigure}
\begin{subfigure}[c]{\linewidth}
\centering
\includegraphics[width=0.85\linewidth]{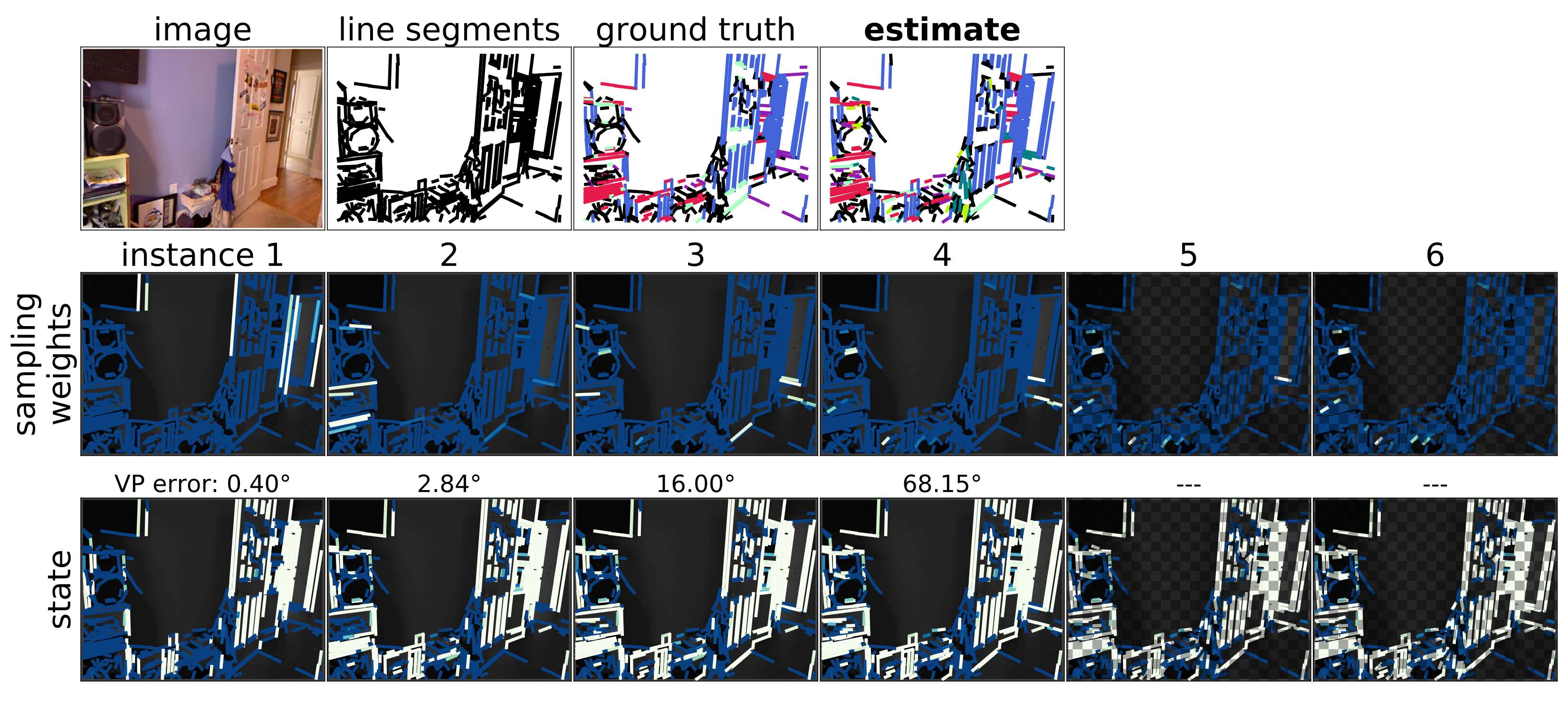}
\end{subfigure}
\vspace{-0.2em}
\caption{Three qualitative examples for VP estimation with CONSAC on our NYU-VP dataset. For each example we show the original image, extracted line segments, line assignments to ground truth VPs, and to final estimates in the first row. In the second and third row, we visualise the generation of the multi-hypothesis $\hat{\set{M}}$ eventually selected by CONSAC. The second row shows the sampling weights per line segment which were used to generate each hypothesis $\vec{\hat{h}} \in \hat{\set{M}}$. The third row shows the resulting state $\vec{s}$.  (\textcolor{blue}{Blue} = low, white = high.) Between rows two and three, we indicate the individual VP errors. The checkerboard pattern and "---" entries indicate instances for which no ground truth is available. The last example is a failure case, where only two out of four VPs were correctly estimated.
}
\label{fig:nyu_result_examples}
\end{figure*}

\begin{figure*}	
\centering
\vspace{-0.4em}
\begin{subfigure}[c]{\linewidth}
\centering
\includegraphics[width=0.835\linewidth]{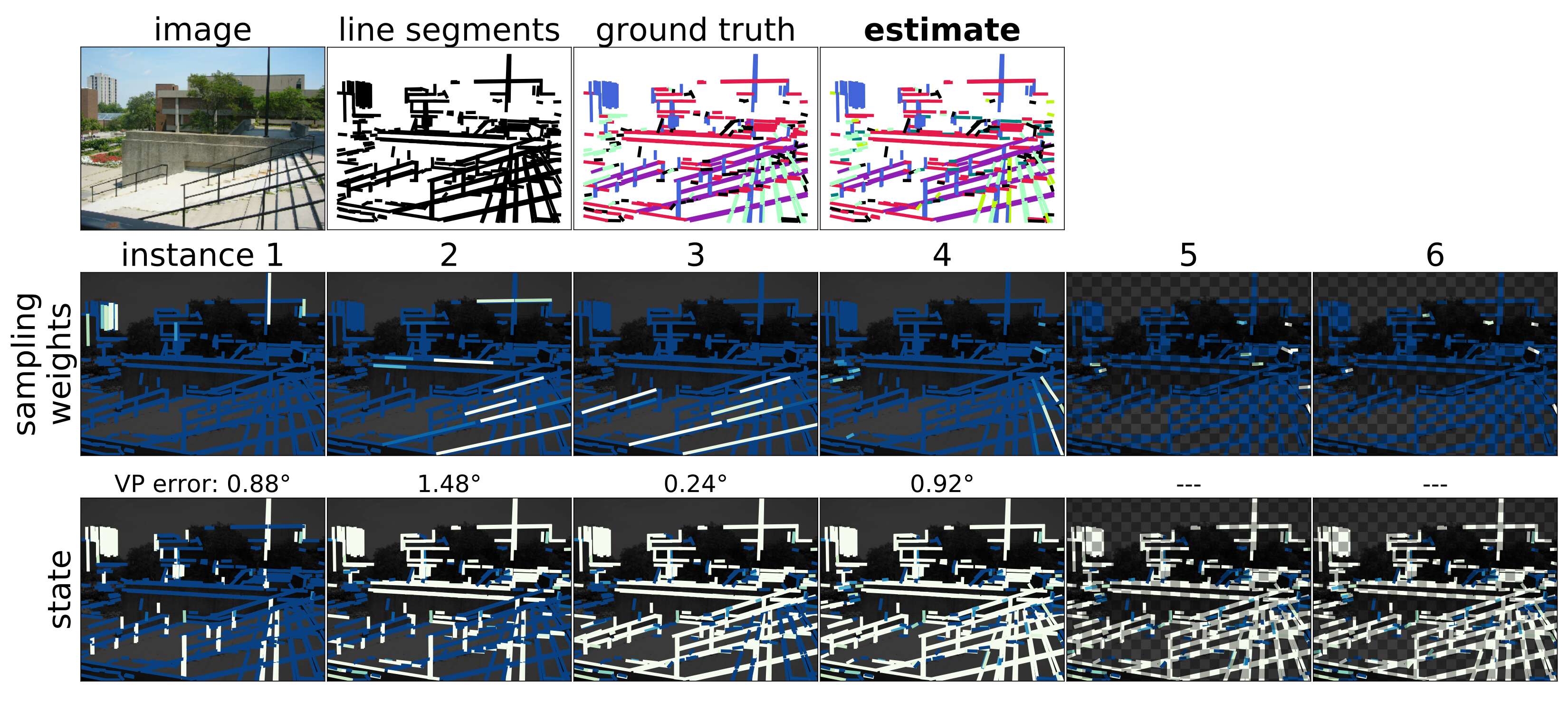}
\end{subfigure}
\begin{subfigure}[c]{\linewidth}
\centering
\includegraphics[width=0.835\linewidth]{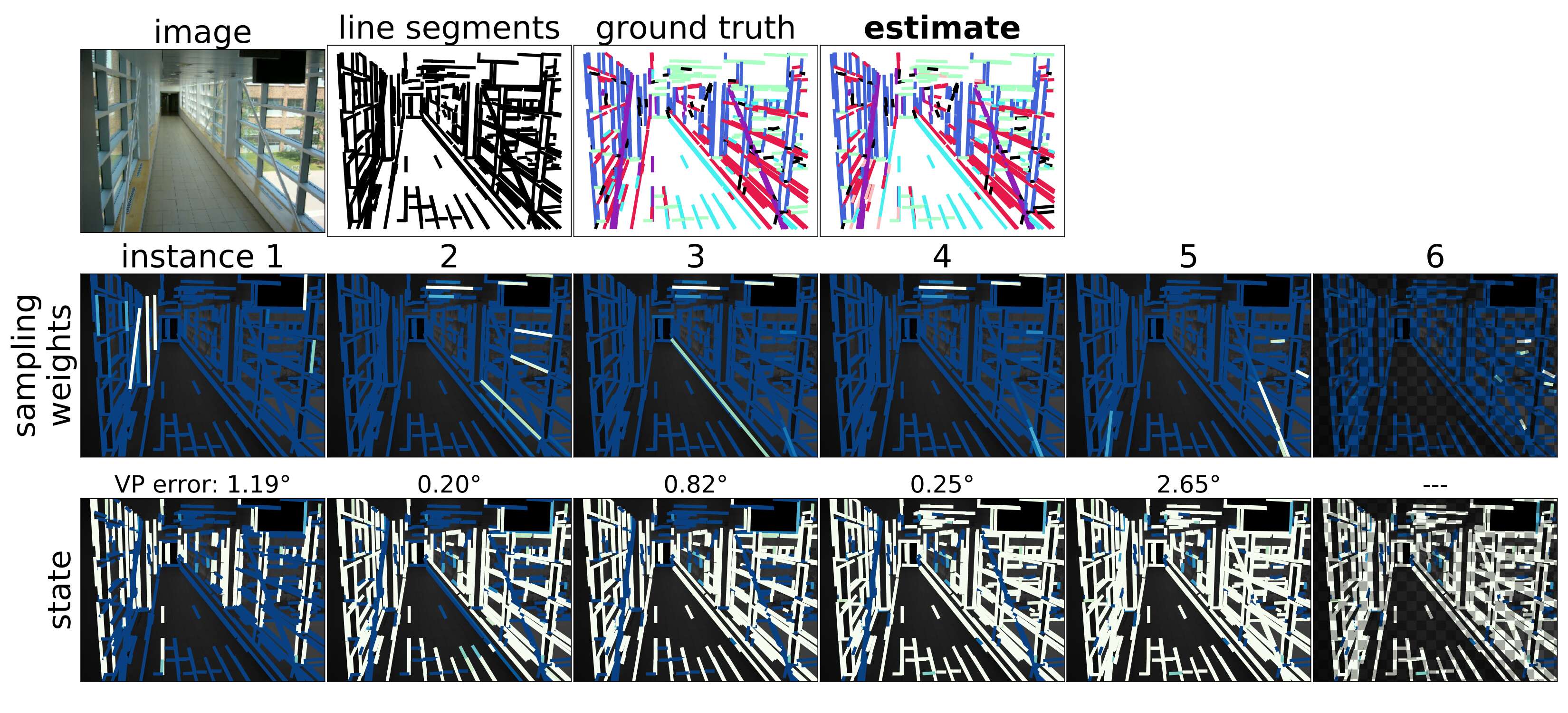}
\end{subfigure}
\begin{subfigure}[c]{\linewidth}
\centering
\includegraphics[width=0.835\linewidth]{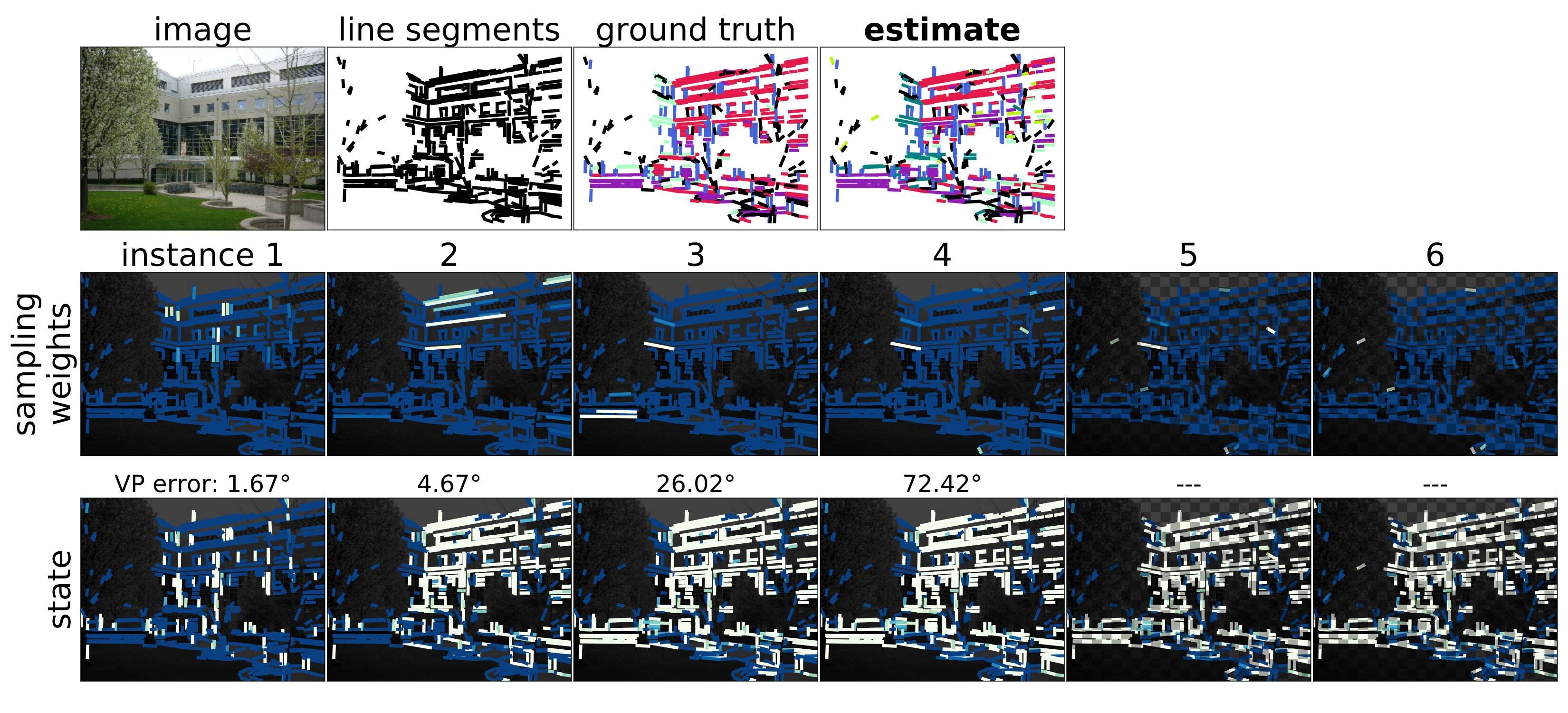}
\end{subfigure}
\vspace{-0.2em}
\caption{Three qualitative examples for VP estimation with CONSAC on the YUD+ dataset. For each example we show the original image, extracted line segments, line assignments to ground truth VPs, and to final estimates in the first row. In the second and third row, we visualise the generation of the multi-hypothesis $\hat{\set{M}}$ eventually selected by CONSAC. The second row shows the sampling weights per line segment which were used to generate each hypothesis $\vec{\hat{h}} \in \hat{\set{M}}$. The third row shows the resulting state $\vec{s}$.  (\textcolor{blue}{Blue} = low, white = high.) Between rows two and three, we indicate the individual VP errors. The checkerboard pattern and "---" entries indicate instances for which no ground truth is available. The last example is a failure case, where only two out of four VPs were correctly estimated.
}
\label{fig:yudplus_result_examples}
\end{figure*}

\paragraph{Results}
For vanishing point estimation, we provide recall curves for errors up to $10^{\circ}$ in Fig.~\ref{fig:vp_recall} for our new NYU-VP dataset, for our YUD+ dataset extension, as well as the original YUD~\cite{denis2008efficient}. We compare CONSAC with the robust multi-model fitting approaches T-Linkage~\cite{magri2014t}, Sequential RANSAC~\cite{vincent2001seqransac}, Multi-X~\cite{barath2018multi}, RPA~\cite{magri2015robust} and RansaCov~\cite{magri2016multiple}, as well as the task-specific vanishing point estimators of Zhai et al.~\cite{zhai2016detecting}, Simon et al.~\cite{simon2018acontrario} and Kluger et al.~\cite{kluger2017deep}. 
We selected the result with the median area under the curve (AUC) of five runs for each method. CONSAC does not find more vanishing points within the $10^{\circ}$ range than state-of-the-art vanishing point estimators, indicated by similar recall values at $10^{\circ}$. However, it does estimate vanishing points more accurately on NYU-VP and YUD+, as the high recall values for low errors ($<4^{\circ}$) show. On YUD~\cite{denis2008efficient}, CONSAC achieves similar or slightly worse recall. Compared to other robust estimators, however, CONSAC performs better than all methods on all datasets across the whole error range.
In Fig.~\ref{fig:nyu_result_examples}, we show additional qualitative results from the NYU-VP dataset, and in Fig.~\ref{fig:yudplus_result_examples}, we show additional qualitative results from the YUD+ dataset.

\subsection{Homography Estimation}
\begin{figure}	
\centering
\includegraphics[width=0.999\linewidth]{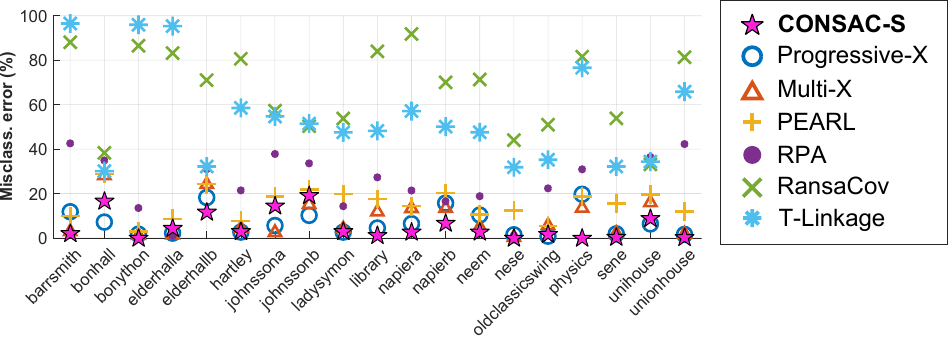}
\caption{\textbf{Homography estimation:} Misclassification errors (in \%, average over five runs) for all homography estimation scenes of AdelaideRMF~\cite{wong2011dynamic}. Graph adapted from~\cite{barath2019progressive}.}
\label{fig:adelaide_per_sequence}
% \vspace{-1em}
\end{figure}  

\begin{figure*}	
\centering
\vspace{-0.4em}
\begin{subfigure}[c]{\linewidth}
\centering
\includegraphics[width=0.85\linewidth]{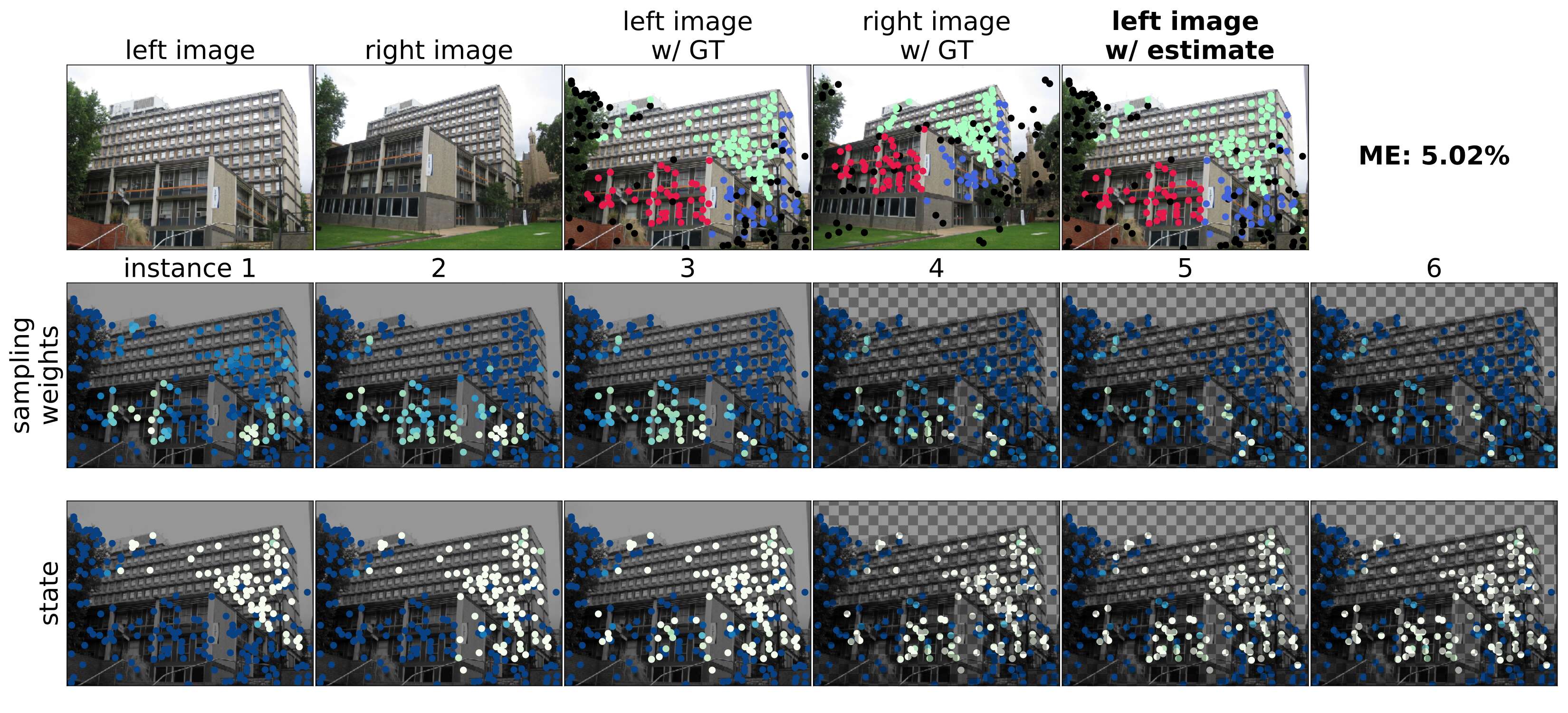}
\end{subfigure}
\begin{subfigure}[c]{\linewidth}
\centering
\includegraphics[width=0.85\linewidth]{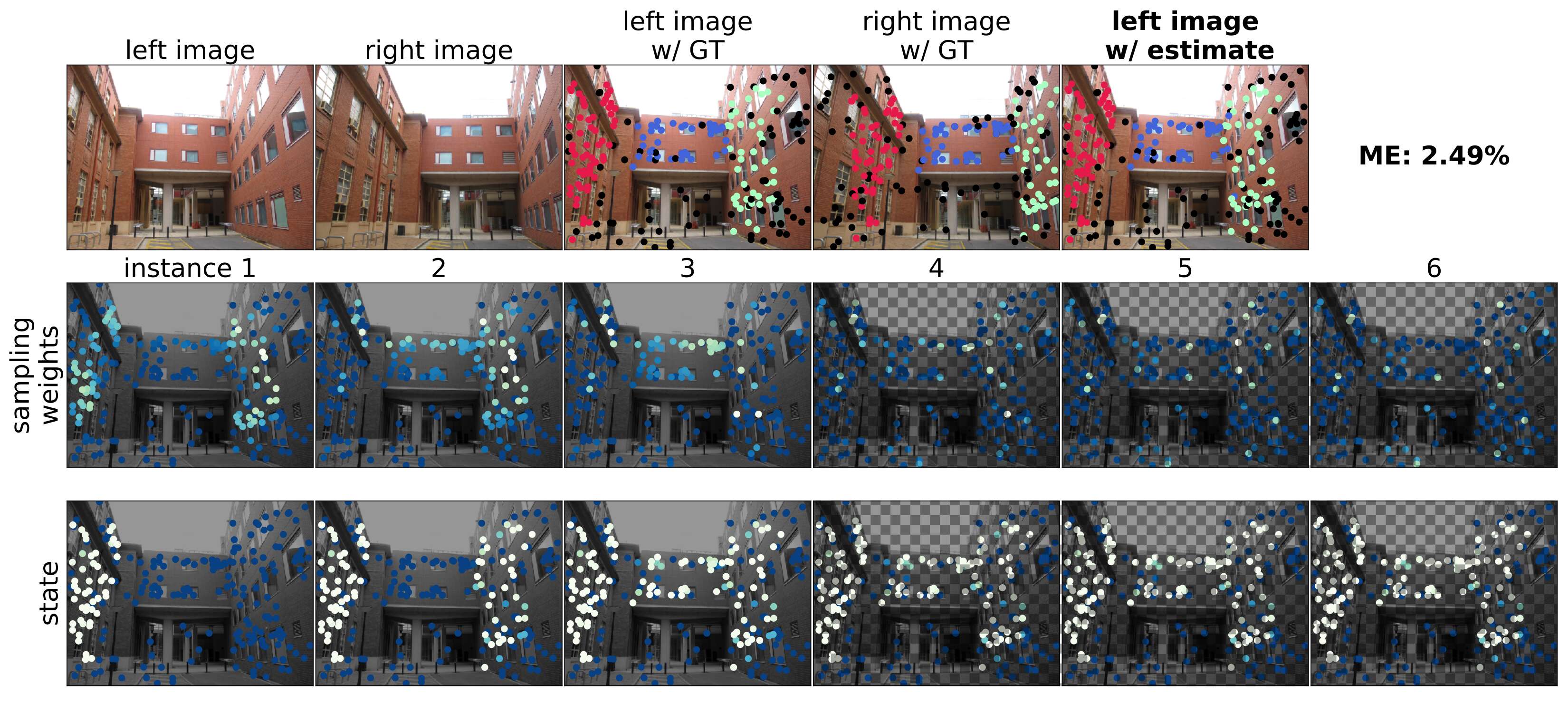}
\end{subfigure}
\begin{subfigure}[c]{\linewidth}
\centering
\includegraphics[width=0.85\linewidth]{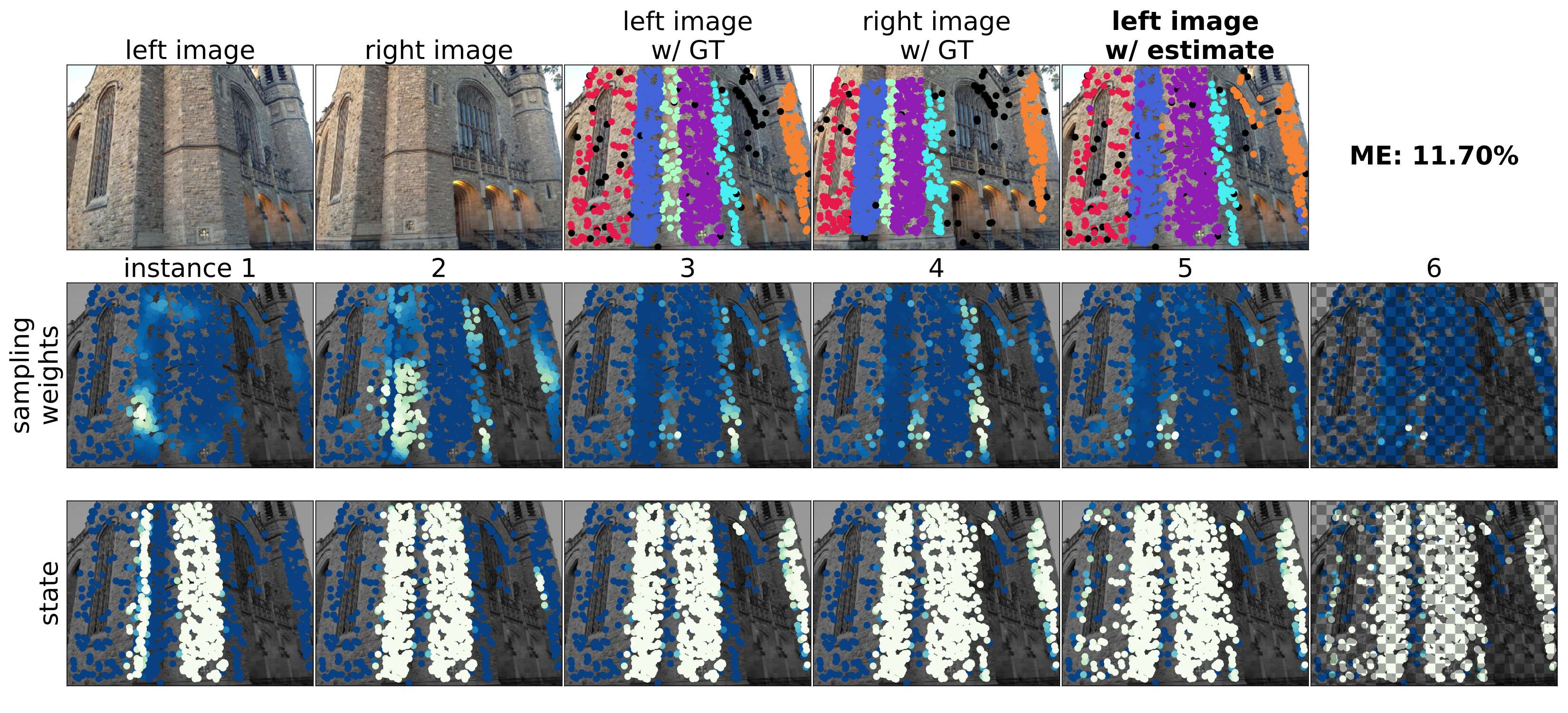}
\end{subfigure}
\caption{Three qualitative examples for homography estimation with CONSAC-S on the AdelaideRMF~\cite{wong2011dynamic} dataset. For each example we show the original images, points with ground truth labels, final estimates, and the misclassification error (ME) in the first row. In the second and third row, we visualise the generation of the multi-hypothesis $\hat{\set{M}}$ eventually selected by CONSAC. The second row shows the sampling weights per point correspondence which were used to generate each hypothesis $\vec{\hat{h}} \in \hat{\set{M}}$. The third row shows the resulting state $\vec{s}$. (\textcolor{blue}{Blue} = low, white = high.) The checkerboard pattern indicates instances which were discarded by CONSAC in the final instance selection step.
}
\label{fig:adelaide_result_examples}
\end{figure*}

\begin{table}
\setlength\tabcolsep{.29em}
    \centering
    \begin{tabu}{l|c|[1.1pt]r|r|r|}
    &  {no. of} & \multirow{2}{*}{CONSAC-S}    & \multicolumn{1}{c|}{\multirow{2}{*}{MCT~\cite{magri2019fitting}}}  & \multicolumn{1}{c|}{Sequential}\\
    &  {planes} &                                   &    & \multicolumn{1}{c|}{RANSAC}\\
    \hline
    barrsmith &         $2$ &   $\mathbf{2.07}$  & $11.29$  & $12.95$ \\
    bonhall &           $6$ &   $\mathbf{16.63}$ & $29.29$ & $20.43$   \\
    bonython &          $1$ &   $\mathbf{0.00}$  & $2.42$  & $\mathbf{0.00} $  \\
    elderhalla &        $2$ &   $\mathbf{4.39}$  & $21.41$  & $16.36 $   \\
    elderhallb &        $3$ &   $\mathbf{11.69}$ & $20.31$ & $18.67$   \\
    hartley &           $2$ &   $\mathbf{2.94}$  & $15.19$  & $9.38$  \\
    johnsona &          $4$ &   $\mathbf{14.48}$ & $18.77$ & $28.04 $   \\
    johnsonb &          $6$ &   $\mathbf{19.17}$ & $33.87$ & $27.46 $   \\
    ladysymon &         $2$ &   $\mathbf{2.95}$  & $16.46$  & $3.80 $   \\
    library &           $2$ &   $\mathbf{1.21}$  & $14.79$  & $11.35 $  \\
    napiera &           $2$ &   $\mathbf{2.72}$  & $21.32$  & $11.66 $   \\
    napierb &           $3$ &   $\mathbf{6.72}$  & $16.83$ & $21.24 $ \\
    neem &              $3$ &   $\mathbf{2.74}$  & $14.36$  & $14.44 $ \\
    nese &              $2$ &   $\mathbf{0.00}$  & $12.83$  & $0.47 $  \\
    oldclass.  &        $2$ &   $1.69$  & $15.20$  & $\mathbf{1.32} $ \\
    physics &           $1$ &   $\mathbf{0.00}$  & $3.21$  & $\mathbf{0.00} $ \\
    sene &              $2$ &   $\mathbf{0.40}$  & $4.80$  & $2.00 $  \\
    unihouse &          $5$ &   $\mathbf{8.84}$  & $34.10$  & $10.69 $ \\
    unionhouse &        $1$ &   $\mathbf{0.30}$  & $1.51$  & $1.51 $ \\
    \hline
    \multicolumn{2}{c|[1.1pt]}{average}  &   $\mathbf{5.21}$ & $16.21$ & $11.14$ \\
    \end{tabu}
    \caption{\textbf{Homography estimation:} Misclassification errors (in \%, average over five runs) for all homography estimation scenes of AdelaideRMF~\cite{wong2011dynamic}.}
    \label{tab:adelaide_detailed_results}
\end{table}
We provide results computed on AdelaideRMF~\cite{wong2011dynamic} for all scenes seperately. In Fig.~\ref{fig:adelaide_per_sequence}, we compare CONSAC-S -- \ie CONSAC trained in a self-supervised manner -- to Progressive-X~\cite{barath2019progressive}, Multi-X~\cite{barath2018multi}, PEARL~\cite{isack2012energy}, RPA~\cite{magri2015robust}, RansaCov~\cite{magri2016multiple} and T-Linkage~\cite{magri2014t}. We adapted the graph directly from~\cite{barath2019progressive}. CONSAC-S achieves state-of-the-art performance on 13 of 19 scenes. Tab.~\ref{tab:adelaide_detailed_results} compares CONSAC-S with MCT~\cite{magri2019fitting} and Sequential RANSAC. We computed results for MCT using code provided by the authors, and used our own implementation for Sequential RANSAC, since no results obtained using the same evaluation protocol (average over five runs) were available in previous works. 
In Fig.~\ref{fig:adelaide_result_examples}, we show additional qualitative results from the AdelaideRMF~\cite{wong2011dynamic} dataset.

\end{appendices}
% \newpage
% \mbox{~}
% \clearpage
%\newpage

{\small
\bibliographystyle{ieee_fullname}
\bibliography{arxiv}
}

\end{document}